\documentclass[lettersize,journal]{IEEEtran}
\usepackage{amsmath,amsfonts}
\usepackage{algorithmic}
\usepackage{algorithm}
\usepackage{array}
\usepackage[caption=false,font=normalsize,labelfont=sf,textfont=sf]{subfig}
\usepackage{textcomp}
\usepackage{stfloats}
\usepackage{url}
\usepackage{verbatim}
\usepackage{graphicx}
\usepackage{cite}
\usepackage{tabularx}
\usepackage{booktabs}
\usepackage{multirow}
\usepackage{adjustbox}
\usepackage{colortbl}
\usepackage{pifont}
\usepackage{makecell}
\definecolor{Gray}{gray}{0.9}

\usepackage{mathtools}
\usepackage{amsmath}
\usepackage{amssymb}
\usepackage{amsthm}
\usepackage{color}
\usepackage{rotating}
\usepackage{bm}
\newcommand{\matx}[1]{{\bm{#1}}} %%Matrix
 \newcommand{\eg}{{\it e.g.}}
 \newcommand{\etal}{{\it et al.}}
\begin{document}

\title{FreeShadow: Training-Free Shadow Removal via Illumination Transfer and Selective Content Preservation in Diffusion Models}

\author{Yinan Wang, Yan Huang, Yong Xu,~\IEEEmembership{Senior Member,~IEEE}, and Patrick Le Callet,~\IEEEmembership{Fellow,~IEEE}
        % <-this % stops a space
\thanks{This work was supported in part by National Key Research and Development Program of China under grant 2024YFE0105400,  in part by Guangdong S\&T programme under grant 2025A0505020016, and in part by National Natural Science Foundation of China under grants 62372186 and 62472179.}
\thanks{Yinan Wang, Yan Huang (\textit{Corresponding author}), and Yong Xu are with the School of Computer Science and Engineering, South China University of Technology, Guangzhou 510006, China (e-mail: csyinan@mail.scut.edu.cn; aihuangy@scut.edu.cn; yxu@scut.edu.cn). Yong Xu is also with the Pazhou Lab, Guangzhou 510005, China.}% <-this % stops a space
\thanks{Patrick Le Callet is with the Polytech Nantes, Université de Nantes, Nantes 44306, France (e-mail: patrick.lecallet@univ-nantes.fr).}
}

% The paper headers
%\markboth{Journal of \LaTeX\ Class Files,~Vol.~14, No.~8, August~2021}%
%{Shell \MakeLowercase{\textit{et al.}}: A Sample Article Using IEEEtran.cls for IEEE Journals}

%\IEEEpubid{0000--0000/00\$00.00~\copyright~2021 IEEE}
% Remember, if you use this you must call \IEEEpubidadjcol in the second
% column for its text to clear the IEEEpubid mark.

\maketitle

\begin{abstract}
Existing supervised and unsupervised shadow removal methods often suffer from limited generalization due to the insufficient diversity of available training datasets, while zero-shot methods tend to produce artifacts and require time-consuming test-time optimization. To address these issues, we propose FreeShadow, a training-free shadow removal method built upon pretrained diffusion models, which exploits diffusion priors for shadow removal without any training or optimization. For illumination recovery, we propose an illumination transfer attention (ITA), which re-weights the self-attention maps in diffusion model to transfer illumination cues from non-shadow to shadow regions. For content preservation, we analyze the effects of illumination variations on self-attention maps and latent high-frequency features in diffusion model, and selectively preserve illumination-invariant components to maintain content fidelity while suppressing residual shadows. We further propose local texture-preserving relighting (LTPR) to mitigate local texture misalignment caused by VAE compression. Extensive experiments demonstrate that our method achieves strong generalization and produces realistic shadow-free images.
\end{abstract}


\section{Introduction}
Shadows, caused by the partial occlusion of light sources, commonly appear in diverse natural scenes. They not only degrade visual quality~\cite{yoon2024generative}, but also hinder the performance of various downstream vision tasks, including image classification~\cite{Zhong_2022_CVPR}, object detection~\cite{sultana2020unsupervised, shao2021hyper}, image segmentation~\cite{zhang2019decoupled}, and image editing~\cite{xie2023consistency}. Therefore, effective shadow removal is essential for improving visual quality and alleviating performance degradation in these tasks.

Recently, deep learning-based shadow removal techniques have demonstrated remarkable performance and have become the dominant paradigm in this domain. These methods can typically be categorized into supervised and unsupervised paradigms. Most existing works~\cite{niu2022boundary, liu2023decoupled, Guo_Huang_Liu_Cheng_Wen_2023, liu2024recasting, xu2025omnisr} are trained in a fully supervised manner, relying on large-scale datasets~\cite{le2019shadow, qu2017deshadownet} composed of paired shadow and shadow-free images. However, acquiring such paired data is labor-intensive and typically spans only a limited range of scene categories. Additionally, inconsistencies in color, illumination, and camera viewpoints may render the pairs unreliable~\cite{hu2019mask, jiang2023learning}, thereby compromising the reliability and generalizability of these supervised models.

To address these limitations, several studies~\cite{hu2019mask, liu2021shadow, zheng2025shadow, jin2021dc} have explored unsupervised shadow removal by learning from unpaired shadow and shadow-free images. Nevertheless, these methods still require a large number of diverse shadow images for training. In addition, when the distribution gap between the shadow and shadow-free domains becomes large, the learned mapping may become unstable~\cite{le2020shadow}. Self-ShadowGAN~\cite{jiang2023learning} further eliminates the requirement for training data by introducing a zero-shot shadow removal method. However, it relies on unstable adversarial optimization, which can easily produce unnatural artifacts. Meanwhile, the network requires multi-step optimization during inference, resulting in relatively long inference time.

\begin{figure}[!t]
    \centering
    \includegraphics[width=\linewidth]{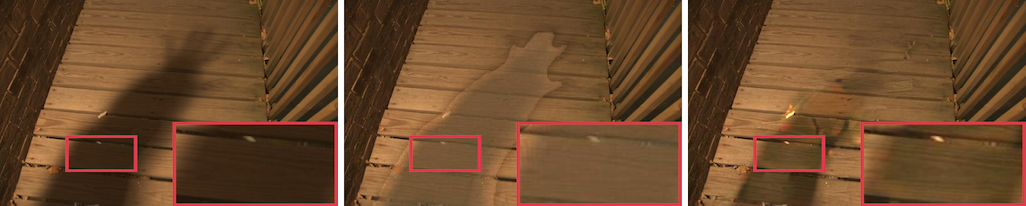}
    \\
    \footnotesize
    \begin{tabularx}{\linewidth}{*{3}{>{\centering\arraybackslash}X}}
        Input & G2R-SNet~\cite{liu2021shadow} (UL) & \mbox{Self-SGAN~\cite{jiang2023learning} (ZS)}
    \end{tabularx}
    \includegraphics[width=\linewidth]{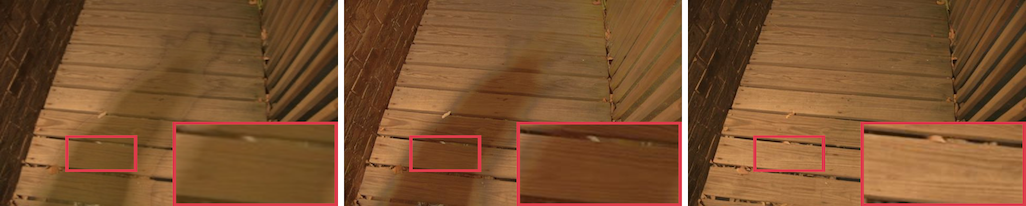}
    \\
    \footnotesize
    \begin{tabularx}{\linewidth}{*{3}{>{\centering\arraybackslash}X}}
        \mbox{\hspace{-0.em}ShadowDiff.~\cite{guo2023shadowdiffusion} (SL)} & \mbox{\hspace{-0.em}StableShadow~\cite{xu2025detail} (SL)} & FreeShadow (TF)
    \end{tabularx}

    \caption{Comparison of shadow removal results on out-of-distribution images. Existing supervised (SL) and unsupervised (UL) methods exhibit limited generalization when applied to unseen domains. Existing zero-shot (ZS) approaches perform poorly in complex scenarios and are prone to producing artifacts. In contrast, our training-free (TF) method exhibits strong generalization and consistently produces visually natural shadow-free images.}
    \label{fig:intro_1}
\end{figure}

Recently, generative diffusion models~\cite{ho2020denoising}, particularly large-scale pretrained text-to-image (T2I) models~\cite{rombach2022high}, have achieved impressive performance across a variety of downstream tasks~\cite{wang2024exploiting, huang2025zero}. Trained on billions of image-text pairs, these T2I models encode rich natural image priors, which can be effectively leveraged to enhance both the generalization and visual realism of shadow removal results. A few recent works have explored the use of pretrained diffusion models for shadow removal. BCDiff~\cite{guo2023boundary} trains a shadow-invariant intrinsic decomposition network and integrates it with a pretrained diffusion model. StableShadow~\cite{xu2025detail} fine-tunes a pretrained diffusion model for the shadow removal task. However, these methods still require training on limited shadow datasets, which inherently restricts their generalizability.

To address the scarcity of shadow removal datasets and fully exploit the rich visual priors of pretrained diffusion models, we propose FreeShadow, a training-free shadow removal framework built upon Stable Diffusion (SD). FreeShadow focuses on two key challenges in shadow removal: illumination restoration and content preservation. For illumination restoration, non-shadow regions provide valuable references for recovering shadow illumination. Accordingly, we introduce illumination transfer attention (ITA), which enhances the attention from shadow-region tokens to non-shadow-region tokens, promoting the transfer of illumination cues from well-lit areas to shadow regions without any training. For content preservation, we propose selective attention map and high-frequency injection (SAMI \& SHFI). Although SD self-attention maps and latent high-frequency features encode structural layouts and fine-grained details, they are also affected by shadow-induced illumination variations. Directly preserving them from the shadow image may improve structural consistency but can introduce residual shadows. Therefore, we analyze the influence of shadows on these representations and selectively inject illumination-insensitive components to preserve content while suppressing shadow residues. Furthermore, to alleviate local texture inconsistencies caused by VAE compression, we design a local texture-preserving relighting (LTPR) module. LTPR combines the local illumination statistics of the SD output with the local texture information of the input image, improving texture fidelity after relighting. 

The main contributions of this work can be summarized as follows: 
\begin{itemize}
 \item We propose a training-free shadow removal framework, which fully leverages the rich visual priors of pretrained diffusion models without requiring paired training data or test-time optimization. As a result, our method achieves remarkable generalization capability and produces high-quality shadow removal results.
 \item We introduce an ITA, which transfers illumination cues from non-shadow regions to shadow regions, leading to more accurate illumination recovery. 
 \item We systematically analyze the influence of shadows on self-attention maps and latent high-frequency features in SD, and selectively inject illumination-insensitive components, thereby preserving structural layouts and fine-grained details while suppressing residual shadows.
 \item We design an LTPR to alleviate local texture inconsistencies caused by VAE compression, thereby improving texture fidelity after relighting.
\end{itemize}

\section{Related Work}
\label{sec:related}
\subsection{Deep Shadow Removal}
In recent years, the remarkable progress of deep learning has significantly advanced the field of image shadow removal. Owing to the availability of paired training datasets, supervised shadow removal methods~\cite{qu2017deshadownet, le2019shadow, niu2022boundary, liu2023decoupled, Guo_Huang_Liu_Cheng_Wen_2023, liu2024recasting, xiao2024homoformer, xu2025omnisr} have garnered considerable research attention. DeshadowNet~\cite{qu2017deshadownet} employs multi-context feature extraction to predict a matte layer for shadow removal. To alleviate artifacts near shadow boundaries, BA-ShadowNet~\cite{niu2022boundary} introduces a boundary-aware shadow removal framework, where a shadow removal branch and a shadow boundary optimization branch are jointly learned and interactively fused to improve boundary consistency. Different from multi-branch or multi-stage designs, DMTN~\cite{liu2023decoupled} proposes a single-stage decoupled multi-task network that explicitly decomposes features for shadow removal, shadow matte estimation, and shadow image reconstruction, thereby improving illumination recovery in shadow regions while preserving non-shadow fidelity. OmniSR~\cite{xu2025omnisr} further integrates semantic and geometric priors through concatenation and attention mechanisms to address shadows induced by indirect illumination. However, collecting paired shadow data is labor-intensive and often limited to a narrow range of scene categories~\cite{hu2019mask, jiang2023learning}. This restricts both the scalability and generalization ability of supervised approaches in real-world scenarios.

To address the limitations of paired data, several unsupervised approaches~\cite{hu2019mask, liu2021shadow, jin2021dc, zheng2025shadow} have been proposed. Mask-ShadowGAN~\cite{hu2019mask} adopts a CycleGAN-based framework to remove shadows using unpaired shadow and shadow-free images. G2R-SNet~\cite{liu2021shadow} synthesizes pseudo-shadows in shadow-free regions to construct paired training data. IIDShadow~\cite{zheng2025shadow} presents a joint learning framework for shadow removal and intrinsic image decomposition using unpaired data. Nevertheless, these methods still require a large number of diverse shadow images for training. In addition, when the distribution gap between the shadow and shadow-free domains becomes large, the learned mapping may become unstable~\cite{le2020shadow}. 

To eliminate the dependence on training data, Self-SGAN~\cite{jiang2023learning} proposes a zero-shot method that learns shadow removal from a single shadow image. It constrains the relit shadow regions to share similar histogram distributions with non-shadow regions through a histogram-based discriminator. Although Self-SGAN alleviates the requirement for training data, it still relies heavily on adversarial training, which is prone to introducing noticeable visual artifacts in the output. Moreover, it requires multi-step optimization during inference, resulting in relatively long inference time.

\subsection{Diffusion-based Shadow Removal}
Recent advances in generative diffusion models~\cite{ho2020denoising} have demonstrated remarkable capabilities in image synthesis. Inspired by their powerful generative potential, numerous studies have extended diffusion models to image restoration tasks, including super-resolution~\cite{hu2025icdsr}, deblurring~\cite{whang2022deblurring}, and inpainting~\cite{zhang2024mmginpainting}. In the domain of shadow removal, several diffusion-based approaches~\cite{mei2024latent, jin2024des3, luo2025diff} have emerged. ShadowDiffusion~\cite{guo2023shadowdiffusion} integrates image and degradation priors into a unified diffusion framework, progressively refining both the output image and shadow mask to generate shadow-free results. DeS3~\cite{jin2024des3} incorporates an adaptive attention mechanism and a ViT-based similarity loss to guide the reverse diffusion process. RRLNet~\cite{liu2024recasting} decomposes images into reflectance and illumination and utilizes a diffusion model to correct illumination degradation in shadow regions. 
%Diff-Shadow~\cite{luo2025diff} introduces a parallel network architecture within a diffusion framework, leveraging global image context to assist local patch restoration.

Pretrained diffusion models~\cite{ramesh2021zero, rombach2022high} encode rich natural image priors, which have been exploited for shadow removal. BCDiff~\cite{guo2023boundary} trains a shadow-invariant intrinsic decomposition network on synthetic shadow data and combines it with a pretrained diffusion model for shadow removal. StableShadow~\cite{xu2025detail} fine-tunes a pretrained SD model using paired shadow datasets. However, these methods still require training on limited shadow data, which constrains their generalization ability. In contrast, our approach is training-free and does not depend on shadow training data. This design preserves the intrinsic image priors of the pretrained diffusion model and enhances the generalization across diverse scenes.

\section{Method}
\label{sec:method}

\begin{figure*}[t]%
	\centering
	\includegraphics[width=0.9\textwidth]{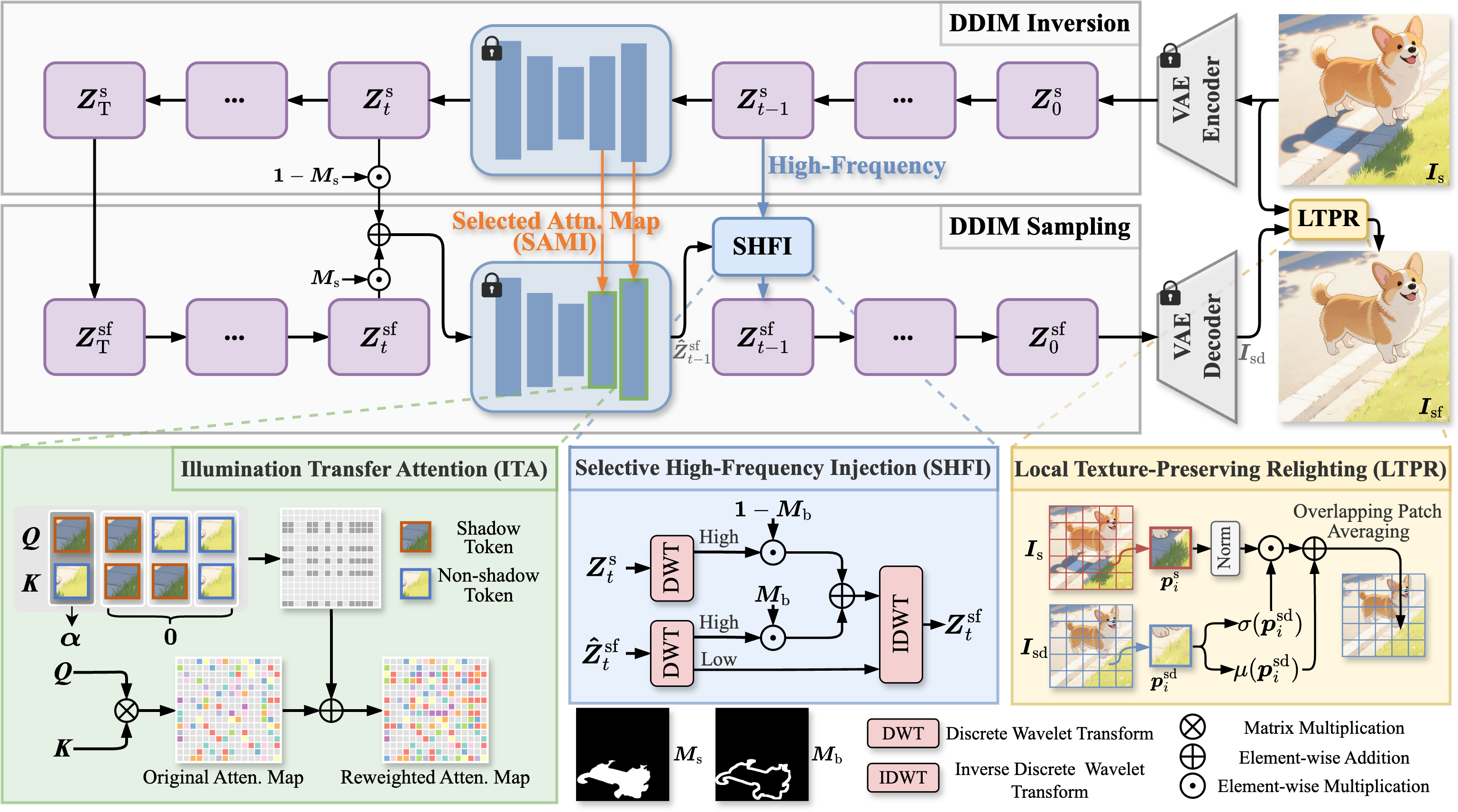}
    \caption{Overall framework of FreeShadow. The shadow image is first inverted into a latent sequence via DDIM inversion. During DDIM sampling, ITA enhances the attention weights from shadow query tokens to non-shadow key tokens to restore illumination. SAMI and SHFI selectively inject shadow-invariant self-attention maps and latent high-frequency features to preserve structures and fine details while suppressing residual shadows. LTPR integrates local illumination statistics from the SD output with shadow image textures to mitigate VAE-induced local texture inconsistencies.}
	\label{architecturefig}
\end{figure*}

\subsection{Overview}
In this work, we address the task of shadow removal by leveraging the powerful generative capabilities of large-scale, pretrained Stable Diffusion (SD) models~\cite{rombach2022high}. As depicted in Fig.~\ref{architecturefig}, FreeShadow comprises two stages: DDIM inversion and DDIM sampling~\cite{songdenoising}. In the inversion stage, the shadow image is mapped into a noisy latent space, producing a sequence of latent variables that encapsulate the visual semantics of the shadow image. Subsequently, the sampling stage starts from the final latent of the inversion process and performs multi-step denoising to generate a shadow-free image. To enable shadow removal during the sampling process, we incorporate several components. Illumination transfer attention (ITA) is introduced to guide illumination recovery. In addition, selective attention map and high-frequency injection (SAMI \& SHFI) are employed to preserve structural integrity and fine-grained details. Furthermore, local texture-preserving relighting (LTPR) is proposed to mitigate local texture misalignment arising from VAE compression.

\subsection{Illumination Transfer Attention}
To facilitate illumination restoration within shadow regions, illumination information from non-shadow areas can serve as a crucial reference. Inspired by this insight, we propose an ITA that modulates the self-attention weights within the denoising network during sampling. Specifically, ITA amplifies the attention weights from shadow tokens (queries) to non-shadow tokens (keys), thereby transferring illumination cues from well-lit areas into shadow regions.

First, the shadow mask $\matx{M}_{\rm{s}}$ is downsampled via max-pooling to match the spatial dimensions of the latent variable. Based on the downsampled shadow mask $\matx{\tilde{M} }_{\rm{s}}$, tokens within the shadow image are categorized into shadow and non-shadow tokens, denoted as 1 and 0, respectively. When the query token corresponds to a shadow region and the key token to a non-shadow region, the attention weight is amplified. In all other cases, the attention remains unchanged. This re-weighting encourages shadow-region tokens to attend more strongly to relevant tokens in well-lit areas, facilitating illumination consistency across regions. The attention re-weighting map is defined as:
\begin{align}
\matx{W}^{(i, j)}=
\left\{
\begin{array}{ll}
\alpha,  & \text {if } \matx{\tilde{M}}^{(i)}_{\rm{s}}=1 \text { and } \matx{\tilde{M}}^{(j)}_{\rm{s}}=0, \\
0, & \text {otherwise},
\end{array}
\right.
\end{align}
where $\alpha$ denotes a positive re-weighting coefficient that controls the strength of illumination transfer. Based on this re-weighting strategy, the formulation of ITA is given by:
\begin{align}
\mathrm{ITA}(\matx{X},\matx{W})=\mathrm{softmax}\left(\frac{\matx{Q}\matx{K}^{\top}}{\sqrt{d}} +\matx{W} \right )\matx{V},
\end{align}
where $\matx{Q}$, $\matx{K}$, and $\matx{V}$ represent the query, key, and value matrices derived from the input feature map $\matx{X}$.

Since ITA relies on accurate illumination guidance from non-shadow regions, it is essential to maintain the fidelity of non-shadow areas throughout the sampling process. To achieve this, at each denoising timestep $t$, the non-shadow regions of the current latent $ \matx{Z}_{t}^{\rm{sf}}$ are replaced with those from the corresponding inversion latent $\matx{Z}_{t}^{\rm{s}}$:
\begin{align}
	\matx{\tilde{Z} }_{t}^{\rm{sf}}= \matx{M}_{\rm{s}} \odot \matx{Z}_{t}^{\rm{sf}}+(1-\matx{M}_{\rm{s}}) \odot \matx{Z}_{t}^{\rm{s}}.
\end{align}
This ensures that the non-shadow regions remain faithful to the original input image. The modified latent $\matx{\tilde{Z}}_{t}^{\rm{sf}}$ is then fed into the denoising network.

\begin{figure*}[t]
    \centering
%    \includegraphics[width=\textwidth]{fig/atten_map/91-5_merged.png}
%    \\[0.1em]
    \includegraphics[width=0.9\textwidth]{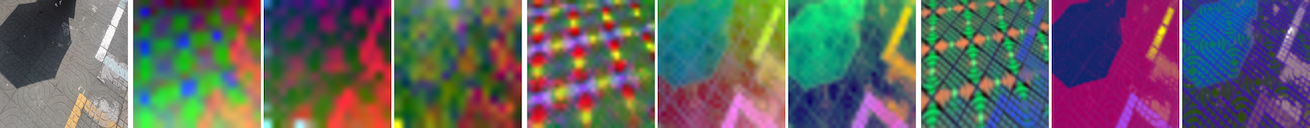}
    \\[0.1em]
    \includegraphics[width=0.9\textwidth]{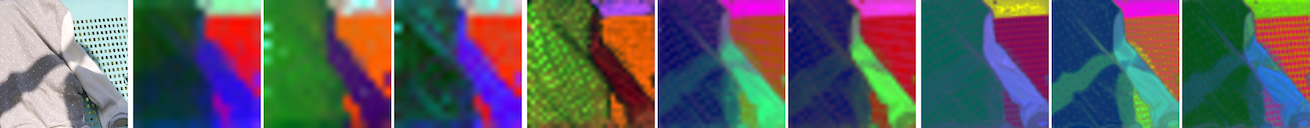}
    \\[0.1em]
%    \includegraphics[width=\textwidth]{fig/atten_map/DSC_0689_merged.png}
%    \\[0.1em]
    \includegraphics[width=0.9\textwidth]{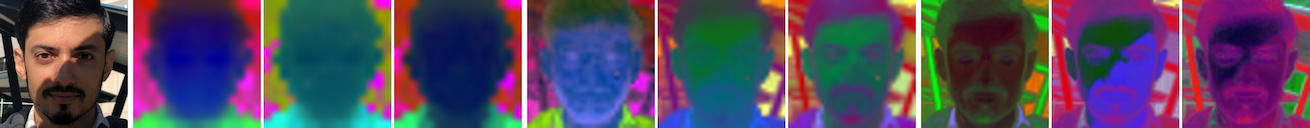}
    \\[0.1em]

    \footnotesize
%    \vspace{-0.4em}
    \begin{tabularx}{0.9\linewidth}{*{10}{>{\centering\arraybackslash}X}}
    	\makecell[c]{Shadow\\[-0.2em] Image} &
    	\makecell[c]{16$\times$16\\[-0.2em] Layer1} &
    	\makecell[c]{16$\times$16\\[-0.2em] Layer2} &
    	\makecell[c]{16$\times$16\\[-0.2em] Layer3} &
    	\makecell[c]{32$\times$32\\[-0.2em] Layer1} &
    	\makecell[c]{32$\times$32\\[-0.2em] Layer2} &
    	\makecell[c]{32$\times$32\\[-0.2em] Layer3} &
    	\makecell[c]{64$\times$64\\[-0.2em] Layer1} &
    	\makecell[c]{64$\times$64\\[-0.2em] Layer2} &
    	\makecell[c]{64$\times$64\\[-0.2em] Layer3}
\end{tabularx}
    
    \caption{Visualization of PCA-reduced self-attention maps across different layers in SD2.1 U-net decoder.}
    \label{fig:atten_map}
\end{figure*}

\begin{figure*}[t]
    \centering
    \includegraphics[width=0.38\textwidth]{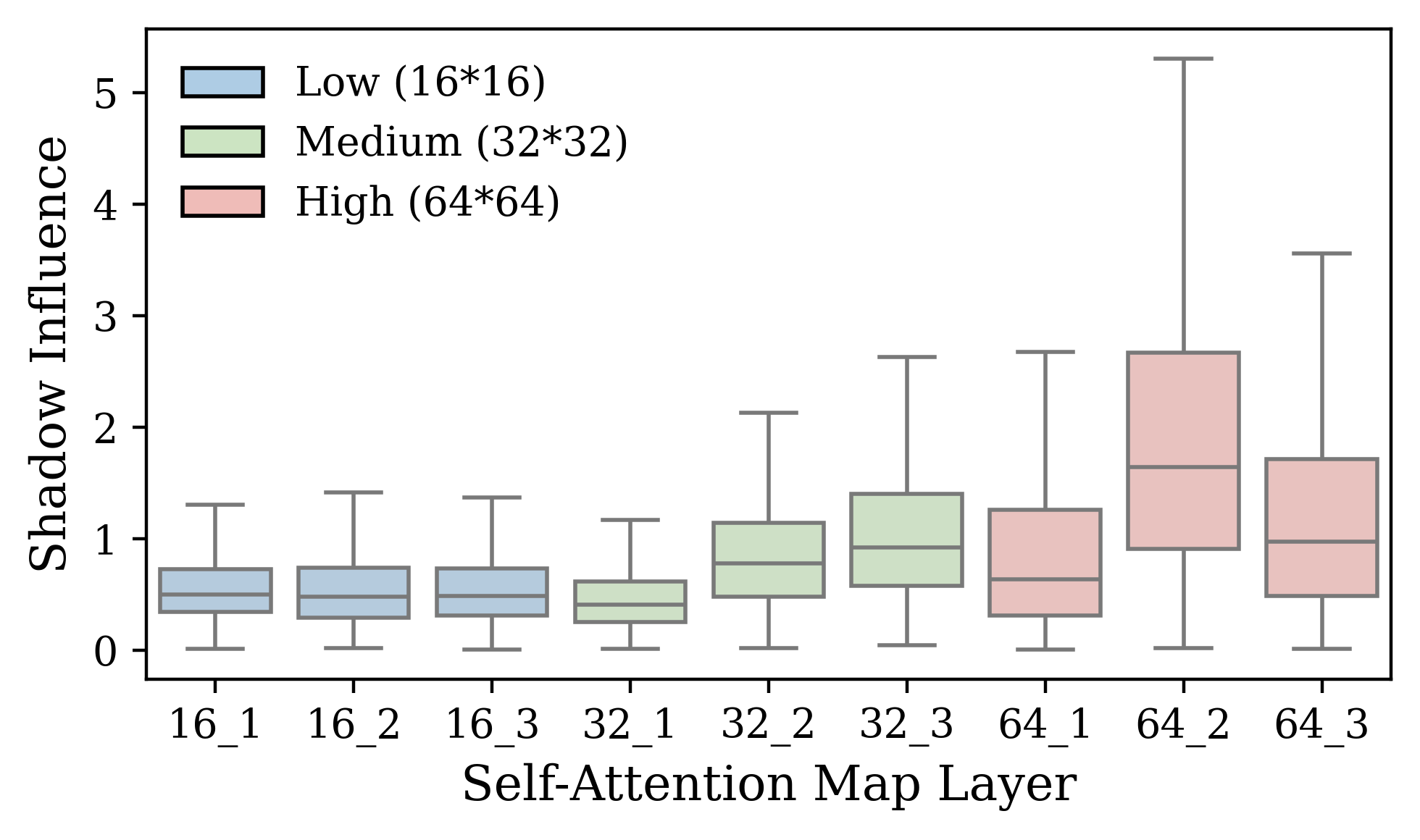}
    \hfil
    \includegraphics[width=0.38\textwidth]{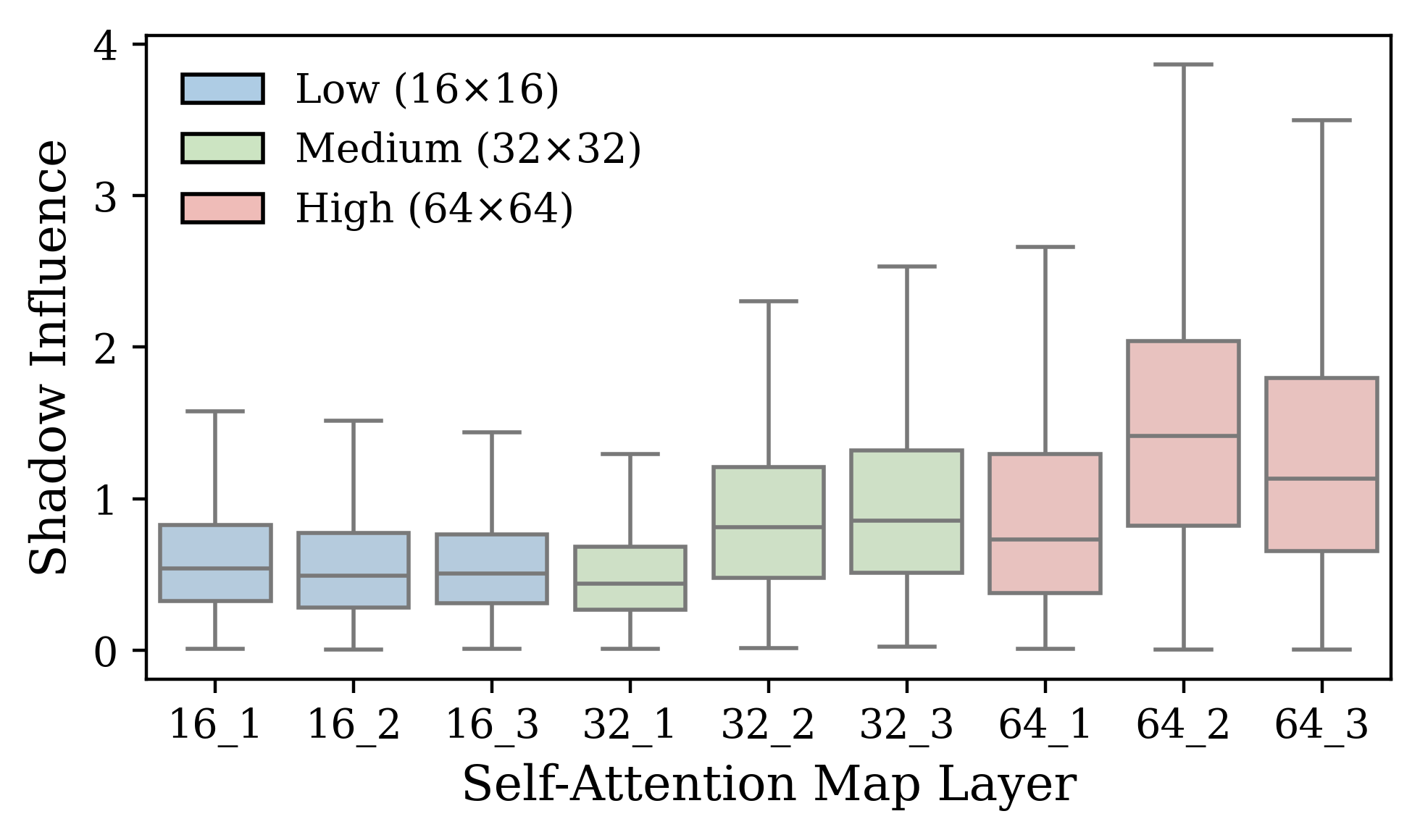}
    \\ \vspace{-0.6em}
    \makebox[0.38\textwidth]{\small (a) ISTD+ Dataset}
    \hfil
    \makebox[0.38\textwidth]{\small (b) SRD Dataset}
    \caption{Boxplot of shadow influence (SI) on SD2.1 self-attention maps of shadow images in the ISTD+ and SRD datasets. Results on more datasets and models are provided in the supplementary material.}
    \label{fig:boxplot}
\end{figure*}

\subsection{Selective Attention Map Injection}
Although ITA effectively restores illumination within shadow regions, it may inadvertently disrupt structural information. As illustrated in Fig.~\ref{fig:atten_map}, the self-attention maps in SD capture the structural cues of images. Therefore, replacing all self-attention maps in the denoising network with those extracted from the inversion process during sampling helps preserve the structural information of the shadow image. However, this operation also leads to residual shadow, as verified in our ablation studies. We further observe that the extent to which self-attention maps are influenced by shadow varies across network layers. Motivated by this finding, we propose a SAMI strategy, which selectively injects self-attention maps that are minimally affected by shadow. The ITA mechanism subsequently performs re-weighting operations over these injected attention maps.

To quantitatively measure the influence of shadows on self-attention maps, we introduce a shadow influence (SI) metric. Specifically, principal component analysis (PCA) is first applied to reduce the dimensionality of the self-attention map from $HW \times HW$ to $HW \times 3$. The SI metric is then computed as the absolute mean difference between the shadow region and its surrounding non-shadow region, formulated as:
\begin{align}
\text{SI}(\matx{A})  = \frac{\left| 
\frac{1}{|\mathcal{S}|} \sum_{i \in \mathcal{S}} \matx{A}_{\text{PCA}}(i) - 
\frac{1}{|\mathcal{N}(\mathcal{S})|} \sum_{j \in \mathcal{N}(\mathcal{S})} \matx{A}_{\text{PCA}}(j) 
\right|}{\frac{1}{HW} \sum_{k=1}^{HW} \matx{A}_{\text{PCA}}(k)},
\end{align}
where $\matx{A}_{\text{PCA}} \in \mathbb{R}^{HW \times 3}$ denotes the PCA-reduced attention map, $\mathcal{S}$ denotes the set of indices within the shadow region, and $\mathcal{N}(\mathcal{S})$ indicates the neighboring non-shadow region. The mask for $\mathcal{N}(\mathcal{S})$ is defined as $\matx{M}_{\mathcal{N}(\mathcal{S})} = (\matx{1} - \matx{\tilde{M}}_{\rm{s}}) \odot \matx{\tilde{M}}_{\rm{b}}$, where $\matx{\tilde{M}}_{\rm{b}}$ denotes the max-pooling downsampled version of shadow boundary mask $\matx{M}_{\rm{b}}$.

Since the self-attention maps within the decoder of SD are less susceptible to noise and more effective at capturing the structural characteristics of images~\cite{tumanyan2023plug}, our analysis is primarily concentrated on the decoder. The SD decoder comprises three resolution levels: low (16$\times$16), medium (32$\times$32), and high (64$\times$64). As illustrated in Fig.~\ref{fig:boxplot}, two key observations can be made: (i) the self-attention maps at the lower-resolution layers are relatively less affected by shadow, which may be attributed to the fact that the lower layers primarily encode semantic information~\cite{tumanyan2023plug} that is inherently less sensitive to illumination variations; and (ii) within the medium- and high-resolution layers, the first self-attention map is generally less influenced by shadow. This can be explained by the observation that the first-layer features capture more information associated with skip connections, which embed rich content cues~\cite{schaerf2025training} that are less impacted by lighting conditions. The visualizations in Fig.~\ref{fig:atten_map} further corroborate this finding. Based on these observations, we selectively inject all three self-attention maps from the low-resolution layer, as well as the first self-attention map from each of the medium- and high-resolution layers.

\begin{figure}[t] \centering
    \includegraphics[width=0.9\linewidth]{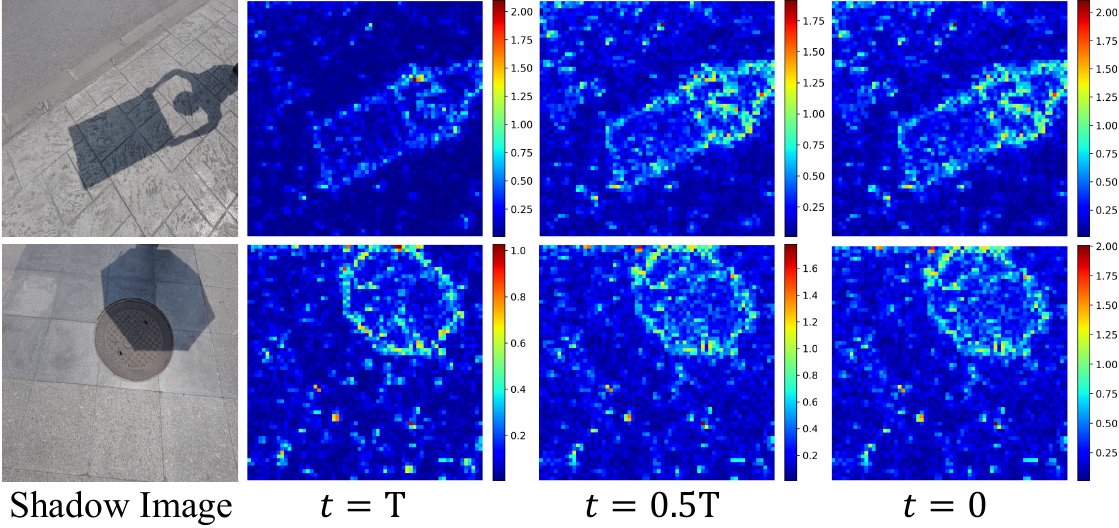}
    \caption{Error map of the high-frequency components in DDIM inversion latent variables between shadow and shadow-free images.} 
    \label{fig:hf_compare}
\end{figure}

\subsection{Selective High-Frequency Injection}
While the injection of self-attention maps facilitates the preservation of the overall structural layout, it remains insufficient for retaining fine-grained details. High-frequency components of latent variables in SD effectively encode fine-level image details~\cite{xu2025stylessp}. To better preserve these details while mitigating residual shadow, we propose a SHFI strategy, which selectively retains the high-frequency components of latent that are minimally affected by shadow.

Due to the abrupt illumination changes at shadow boundaries, fine details in these regions often suffer severe degradation, whereas details in the interior shadow regions typically remain intact. As shown in Fig.~\ref{fig:hf_compare}, the discrepancy between the high-frequency components of the latent for shadow and shadow-free images is predominantly concentrated along the shadow boundaries. To mitigate this, during the sampling process, the high-frequency components of the denoised latent $\matx{\hat{Z}}_{t}^{\rm{sf}}$ in non-boundary regions are replaced with their counterparts from the inverted latent $\matx{Z}_{t}^{\rm{s}}$, while keeping the shadow boundaries unchanged:
\begin{align}
\matx{H}_{t}^{\rm{s}}, \matx{L}_{t}^{\rm{s}} & = \operatorname{DWT}(\matx{Z}_{t}^{\rm{s}}), \\
\hat{\matx{H}}_{t}^{\rm{sf}}, \hat{\matx{L}}_{t}^{\rm{sf}} & = \operatorname{DWT}(\hat{\matx{Z}}_{t}^{\rm{sf}}), \\
\matx{Z}_{t}^{\rm{sf}} = \operatorname{IDWT}(\matx{H}_{t}^{\rm{s}}\odot (\matx{1} &- \matx{\tilde{M}}_{\rm{b}}) + \hat{\matx{H}}_{t}^{\rm{sf}}\odot \matx{\tilde{M}}_{\rm{b}}, \hat{\matx{L}}_{t}^{\rm{sf}}).
\end{align}
Here, $\matx{H}$ and $\matx{L}$ denote the high- and low-frequency components obtained via Discrete Wavelet Transform (DWT). This design preserves reliable detail in regions with minimal degradation while simultaneously exploiting the generative capacity of the pretrained diffusion model to reconstruct the fine details lost along shadow boundaries.

\subsection{Local Texture-Preserving Relighting}
SD employs a Variational Autoencoder (VAE) to map images from the pixel space ($W \times H \times 3$) to a latent space ($\frac{W}{8} \times \frac{H}{8} \times 4$) for the diffusion process. However, this mapping is inherently lossy, leading to local texture misalignments between the shadow removal result and the input shadow image. Since the local mean and variance of an image effectively characterize its illumination distribution while being less affected by fine texture variations, we propose a LTPR strategy. This method injects the local mean and variance of the SD-generated shadow-free image $\matx{I}_{\rm{sd}}$ into the original shadow image $\matx{I}_{\rm{s}}$, thereby fusing the illumination information restored by SD with the original local texture details of the shadow image.

Specifically, sliding window of size $k \times k$ and stride $d$ is used to extract overlapping patches $\{\matx{p}_i^{\rm{s}}\}$ and $\{\matx{p}_i^{\rm{sd}}\}$ from $\matx{I}_{\rm{s}}$ and $\matx{I}_{\rm{sd}}$, respectively. The LTPR patch $\matx{p}_i^{\rm{ltpr}}$ is computed as follows:
\begin{align}
	\matx{p}_i^{\rm{ltpr}} = \frac{\matx{p}_i^{\rm{s}} - \mu(\matx{p}_i^{\rm{s}})}{\sigma(\matx{p}_i^{\rm{s}})} \cdot \sigma(\matx{p}_i^{\rm{sd}}) + \mu(\matx{p}_i^{\rm{sd}}),
\end{align}
where $\mu(\cdot)$ and $\sigma(\cdot)$ denote the channel-wise mean and standard deviation, respectively. The final LTPR result $\matx{I}_{\rm{ltpr}}$ is obtained by reconstructing the image from these LTPR patches and averaging their overlapping regions. Similar to SHFI, we preserve the boundary texture details from the SD-generated image by retaining its boundary regions:
\begin{align}
	\matx{I}_{\rm{sf}}= \matx{M}_{\rm{b}} \odot \matx{I}_{\rm{sd}}+(1-\matx{M}_{\rm{b}}) \odot \matx{I}_{\rm{ltpr}}.
	\label{eq:lsi_blending}
\end{align}

\subsection{Shadow Removal Redirect Guidance}
To enhance the shadow removal performance, shadow removal redirect guidance (SRRG) is incorporated into the sampling process. The final predicted noise $\hat{\epsilon}_{t}$ is computed as:
\begin{align}
	\hat{\epsilon}_{t} & = \epsilon_\theta\left ( \matx{Z}_{t}^{\rm{sf}},t \right ) + s \left ( \epsilon^{\operatorname{SR}}_\theta\left(\matx{Z}_{t}^{\rm{sf}},\, t \right)- \epsilon_\theta\left ( \matx{Z}_{t}^{\rm{sf}},t \right )\right ),
	\label{eq:srgs}
\end{align}
where $\epsilon^{\operatorname{SR}}_\theta$ represents the application of ITA and SAMI within the network to guide the sampling trajectory towards effective shadow removal. $s$ denotes the shadow removal guidance scale, which is empirically set to $7$. Subsequently, the updated noise prediction $\hat{\epsilon}_{t}$ is used to generate the latent $\hat{\matx{Z}}_{t-1}^{\rm{sf}}$ at the next time step.

When $s > 1$, the sampling process is increasingly biased toward the shadow removal direction $\epsilon^{\operatorname{SR}}_\theta\left(\matx{Z}_{t}^{\rm{sf}},\, t \right)$ and away from the original generative direction $\epsilon_\theta\left ( \matx{Z}_{t}^{\rm{sf}},t \right )$. Since the sampling process begins from the DDIM inversion of a shadow image and DDIM is approximately reversible, the original generative direction can be interpreted as a reconstruction path of the shadow input. Through this iterative guidance, the sampling path is progressively redirected, shifting the latent distribution toward the target shadow-free direction, thus achieving shadow removal.

\section{Experiments}
\label{sec:exp}

\subsection{Experimental Settings}
\subsubsection{Implementation Details}
%\noindent
%{\bf Implementation Details.}
Our method is implemented based on the frozen SD 2.1 model~\cite{rombach2022high}, without any parameter updates. We employ DDIM inversion and sampling with $T = 50$ diffusion steps. For ITA, the re-weighting coefficient $\alpha$ is set to $1.5$. The shadow boundary mask $\matx{M}_{\rm{b}}$ is defined as $\matx{M}_{\rm{b}} = \operatorname{Dilation}(\matx{M}_{\rm{s}}) - \operatorname{Erosion}(\matx{M}_{\rm{s}})$, with both dilation and erosion operations using a kernel size of $19$. For LTPR, the sliding window size $k$ and stride $d$ are set to $7$ and $1$, respectively. The pseudocode and extensions of our method to other base models (\eg, SDXL~\cite{podell2023sdxl}) are provided in the supplementary material.

\subsubsection{Datasets}
%\noindent
%{\bf Datasets.}
The performance evaluation is conducted on three widely used general shadow removal datasets, namely ISTD+~\cite{le2019shadow}, UIUC+~\cite{guo2012paired}, and SRD~\cite{qu2017deshadownet}, as well as a portrait shadow removal dataset, PSM~\cite{zhang2020portrait}. To mitigate illumination discrepancies between shadow and shadow-free images, we follow the same procedure used in ISTD+ to adjust the illumination of the ground-truth shadow-free images in UIUC+. To evaluate the generalization ability of shadow removal models, we adopt the evaluation protocol in~\cite{guo2024single, xu2025detail}. Specifically, training-based methods are first trained on ISTD+ and then tested on UIUC+, SRD, and PSM.

\begin{table*}[!t]
\centering
\caption{Quantitative comparisons on the ISTD+, UIUC+ and SRD datasets. Yellow, green, and red highlights indicate the best-performing method among the supervised learning (\colorbox[rgb]{ 1,  .976,  .835}{SL}), unsupervised learning (\colorbox[rgb]{ .867,  .949,  .831}{UL}), and the zero-shot (\colorbox[rgb]{ .973,  .875,  .871}{ZS}) methods. ``-'' are absent results due to publicly unavailable code.}
\footnotesize
\renewcommand{\arraystretch}{0.9}
%\adjustbox{width=0.95\textwidth}{
    \begin{tabular}{c l c c c c c c c c c c}
        \toprule
        \multirow{3}{*}{\textbf{Type}} & \multirow{3}{*}{\textbf{Method}} & \multirow{3}{*}{\textbf{Venue}} & \multicolumn{3}{c}{\textbf{ISTD+ Dataset}} & \multicolumn{3}{c}{\textbf{UIUC+ Dataset}} & \multicolumn{3}{c}{\textbf{SRD Dataset}} \\ 
        \cmidrule(lr){4-6}\cmidrule(lr){7-9}\cmidrule(lr){10-12}
        & & & {MAE$\downarrow$} & {SSIM$\uparrow$} & {PSNR$\uparrow$} & {MAE$\downarrow$} & {SSIM$\uparrow$} & {PSNR$\uparrow$} & {MAE$\downarrow$} & {SSIM$\uparrow$} & {PSNR$\uparrow$}\\
%       	\midrule
%            & Input Image & -     & 36.95  & 2.40  & 8.40 \\
        \midrule
%        \multirow{7}{*}{SL}
        \multirow{7}{*}{{SL}}
        & DMTN~\cite{liu2023decoupled} & TMM'23 & 3.09  & 0.972  & 33.68  & 10.50  & 0.877  & 24.02  & 10.57  & 0.896  & 22.69  \\
      & ShadowDiffusion~\cite{guo2023shadowdiffusion} & CVPR'23 & 2.72  & \cellcolor[rgb]{ 1,  .976,  .835}0.975  & \cellcolor[rgb]{ 1,  .976,  .835}35.67  & \cellcolor[rgb]{ 1,  .976,  .835}7.52  & \cellcolor[rgb]{ 1,  .976,  .835}0.897  & 27.05  & 7.83  & 0.913  & 25.65  \\
      & DeS3~\cite{jin2024des3} & AAAI'24 & 3.86  & 0.957  & 31.39  & 11.74  & 0.862  & 23.29  & 10.45  & 0.898  & 22.40  \\
      & HomoFormer~\cite{xiao2024homoformer} & CVPR'24 & \cellcolor[rgb]{ 1,  .976,  .835}2.64  & \cellcolor[rgb]{ 1,  .976,  .835}0.975  & 35.35  & 7.58  & 0.891  & \cellcolor[rgb]{ 1,  .976,  .835}27.21  & \cellcolor[rgb]{ 1,  .976,  .835}7.21  & \cellcolor[rgb]{ 1,  .976,  .835}0.928  & \cellcolor[rgb]{ 1,  .976,  .835}26.05  \\
      & MSRDNet~\cite{huang2025image} & PR'25 & 2.90  & 0.972  & 34.94  & 7.82  & 0.890  & 26.55  & 7.78  & 0.917  & 25.59  \\
      & OmniSR~\cite{xu2025omnisr} & AAAI'25 & 3.08  & 0.970  & 33.35  & 7.65  & 0.891  & 26.16  & 7.85  & 0.920  & 26.03  \\
      & StableShadow~\cite{xu2025detail} & CVPR'25 & 2.89  & 0.974  & 35.19  & 9.46  & 0.884  & 25.92  & 8.60  & 0.918  & 25.54  \\
		\midrule
%		\multirow{7}{*}{UL}
		\multirow{5}{*}{{UL}}
        & Mask-ShadowGAN~\cite{hu2019mask} & ICCV'19 & 4.93  & 0.939  & 28.34  & 10.88  & 0.872  & 23.92  & 11.55  & 0.875  & 22.21  \\
      & DC-ShadowNet~\cite{jin2021dc} & ICCV'21 & 7.77  & 0.926  & 25.03  & 11.16  & 0.870  & 23.49  & 11.80  & 0.865  & 21.79  \\
      & G2R-SNet~\cite{liu2021shadow} & CVPR'21 & 3.86  & 0.944  & 30.52  & \cellcolor[rgb]{ .867,  .949,  .831}8.46  & \cellcolor[rgb]{ .867,  .949,  .831}0.874  & \cellcolor[rgb]{ .867,  .949,  .831}25.80  & \cellcolor[rgb]{ .867,  .949,  .831}8.40  & \cellcolor[rgb]{ .867,  .949,  .831}0.899  & \cellcolor[rgb]{ .867,  .949,  .831}24.23  \\
      & BCDiff~\cite{guo2023boundary} & ICCV'23 & 3.47  & \cellcolor[rgb]{ .867,  .949,  .831}0.959  & \cellcolor[rgb]{ .867,  .949,  .831}32.11  & -     & -     & -     & -     & -     & - \\
      & IIDSR~\cite{zheng2025shadow} & AAAI'25 & \cellcolor[rgb]{ .867,  .949,  .831}3.21  & -     & -     & -     & -     & -     & -     & -     & - \\
        \midrule
%		\multirow{7}{*}{TF}
		\multirow{6}{*}{{ZS}}
        & Yang~\etal~\cite{yang2012shadow} & TIP'12 & 15.90  & 0.706  & 20.26  & -     & -     & -     & -     & -     & - \\
      & Guo~\etal~\cite{guo2012paired} & TPAMI'12 & 6.09  & 0.925  & 25.52  & -     & -     & -     & -     & -     & - \\
      & Self-SGAN~\cite{jiang2023learning} & IJCV'23 & 4.33  & 0.941  & 29.44  & 7.55  & 0.882  & 26.37  & 8.99  & 0.902  & 24.49  \\
      & FLUX-Kontext~\cite{labs2025flux} & arXiv'25  & 7.08  & 0.913  & 26.89  & 13.44  & 0.827  & 22.43  & 13.03  & 0.854  & 21.89  \\
      & Qwen-Image-Edit~\cite{wu2025qwen} & arXiv'25  & 19.28  & 0.536  & 14.81  & 19.54  & 0.750  & 19.44  & 23.85  & 0.567  & 15.93  \\
      & Ours  & -     & \cellcolor[rgb]{ .973,  .875,  .871}2.90  & \cellcolor[rgb]{ .973,  .875,  .871}0.960  & \cellcolor[rgb]{ .973,  .875,  .871}33.80  & \cellcolor[rgb]{ .973,  .875,  .871}7.15  & \cellcolor[rgb]{ .973,  .875,  .871}0.898  & \cellcolor[rgb]{ .973,  .875,  .871}27.88  & \cellcolor[rgb]{ .973,  .875,  .871}6.15  & \cellcolor[rgb]{ .973,  .875,  .871}0.936  & \cellcolor[rgb]{ .973,  .875,  .871}28.47  \\
        \bottomrule
    \end{tabular}
%}
%\vspace{-0.01cm}
\label{tab:aistd_res}
\end{table*}

\subsection{Comparison with State-of-the-Art Methods}
We compare our method with three categories of state-of-the-art methods: (i) supervised methods (DMTN~\cite{liu2023decoupled}, ShadowDiffusion~\cite{guo2023shadowdiffusion}, DeS3~\cite{jin2024des3}, HomoFormer~\cite{xiao2024homoformer}, MSRDNet~\cite{huang2025image}, OmniSR~\cite{xu2025omnisr} and StableShadow~\cite{xu2025detail}), (ii) unsupervised methods (Mask-ShadowGAN~\cite{hu2019mask}, DC-ShadowNet~\cite{jin2021dc}, G2R-SNet~\cite{liu2021shadow}, BCDiff~\cite{guo2023boundary} and IIDShadow~\cite{zheng2025shadow}), and (iii) zero-shot methods (Yang~\etal~\cite{yang2012shadow}, Guo~\etal~\cite{guo2012paired}, Self-SGAN~\cite{jiang2023learning}, FLUX-Kontext~\cite{labs2025flux} and Qwen-Image-Edit~\cite{wu2025qwen}). For text-guided image editing models (\eg, FLUX-Kontext and Qwen-Image-Edit), the prompt ``Remove all shadows from the image'' is employed. For fair comparison, we produce their results using publicly available implementations and pretrained models provided by the respective authors.

\begin{table}[!t]
\centering
\caption{Quantitative comparisons on the PSM dataset.}
\footnotesize
\setlength{\tabcolsep}{0.5em}
% \vspace{-0.1cm}
\renewcommand{\arraystretch}{0.9}
%\adjustbox{width=1\linewidth}{
    \begin{tabular}{cccccc}
        \toprule
        	Type & Method & Venue & MAE$\downarrow$ & SSIM$\uparrow$ & PSNR$\uparrow$ \\
        \midrule
        \multirow{7}{*}{SL}
        & DMTN~\cite{liu2023decoupled} & TMM'23 & 17.27  & 0.797  & 18.85  \\
      & ShadowDiffusion~\cite{guo2023shadowdiffusion} & CVPR'23 & 10.55  & 0.863  & 22.28  \\
      & DeS3~\cite{jin2024des3} & AAAI'24 & 18.97  & 0.779  & 17.72  \\
      & HomoFormer~\cite{xiao2024homoformer} & CVPR'24 & 11.05  & 0.862  & 21.46  \\
      & MSRDNet~\cite{huang2025image} & PR'25 & \cellcolor[rgb]{ 1,  .976,  .835}10.21  & \cellcolor[rgb]{ 1,  .976,  .835}0.876  & \cellcolor[rgb]{ 1,  .976,  .835}22.35  \\
      & OmniSR~\cite{xu2025omnisr} & AAAI'25 & 14.99  & 0.809  & 19.35  \\
      & StableShadow~\cite{xu2025detail} & CVPR'25 & 17.74  & 0.806  & 18.51  \\  
		\midrule
		\multirow{3}{*}{UL}
		& Mask-ShadowGAN~\cite{hu2019mask} & ICCV'19 & 22.14  & 0.756  & 16.55  \\
      & DC-ShadowNet~\cite{jin2021dc} & ICCV'21 & 22.39  & 0.733  & 16.63  \\
      & G2R-SNet~\cite{liu2021shadow} & CVPR'21 & \cellcolor[rgb]{ .867,  .949,  .831}10.27  & \cellcolor[rgb]{ .867,  .949,  .831}0.866  & \cellcolor[rgb]{ .867,  .949,  .831}21.85  \\
        \midrule
		\multirow{4}{*}{ZS}
		& Self-SGAN~\cite{jiang2023learning} & IJCV'23 & 12.42  & 0.849  & 19.55  \\
      & FLUX-Kontext~\cite{labs2025flux} & arXiv'25 & 13.97  & 0.832  & 20.29  \\
      & Qwen-Image-Edit~\cite{wu2025qwen} & arXiv'25 & 30.37  & 0.555  & 13.41  \\
      & Ours  & -     & \cellcolor[rgb]{ .973,  .875,  .871}7.31  & \cellcolor[rgb]{ .973,  .875,  .871}0.903  & \cellcolor[rgb]{ .973,  .875,  .871}24.44  \\
        \bottomrule
    \end{tabular}
%}
	\label{tab:psm}
\end{table}

\subsubsection{Quantitative Comparison}
%\noindent
%{\bf Quantitative Comparison.}
Table~\ref{tab:aistd_res} presents the quantitative comparisons on the ISTD+, UIUC+, and SRD datasets. On the ISTD+ dataset, our method outperforms all existing unsupervised and training-free approaches. The performance gap between our method and the best-performing supervised models can be attributed to the fact that supervised methods typically exploit paired training datasets to capture distributional biases. Under the cross-dataset evaluation setting on UIUC+ and SRD, our approach achieves the best performance among all compared supervised, unsupervised, and zero-shot methods. This demonstrates the strong generalization ability of our method. To further evaluate the robustness of different methods on portrait shadow removal, Table~\ref{tab:psm} reports the quantitative results on the PSM dataset. Compared with existing methods, our method achieves the best performance across all metrics. These results indicate that our method is not limited to general-scene shadow removal, but also generalizes well to portrait shadow removal.

We further compare the computational complexity of different methods in Table~\ref{tab:compu}. The runtime is measured on $512 \times 512$ input images using a single RTX 4090D GPU. Since our method leverages a pretrained diffusion model and does not require task-specific training for shadow removal, its runtime is higher than some supervised methods. Nevertheless, our method achieves state-of-the-art performance with only 20 diffusion steps. Moreover, the inference time of our method with 20 steps is significantly lower than other zero-shot methods, requiring only 4.5\% of the runtime of Self-SGAN. These results demonstrate that our method achieves a practical trade-off between restoration accuracy and computational efficiency.

\begin{table}[!t]
\centering
\caption{Comparison of computational complexity and performance on the UIUC+ Dataset.}
\footnotesize
\setlength{\tabcolsep}{0.4em}
% \vspace{-0.1cm}
\renewcommand{\arraystretch}{0.9}
%\adjustbox{width=1\linewidth}{
    \begin{tabular}{ccccc}
        \toprule
        	Type & Method & MAE$\downarrow$ & Params. (M)$\downarrow$ & Runtime (s)$\downarrow$ \\
        \midrule
        \multirow{5}{*}{SL}
        & ShadowDiffusion~\cite{guo2023shadowdiffusion} & 7.52  & 60.7  & 0.59  \\
      & DeS3~\cite{jin2024des3} & 11.74  & 108.6  & 30.60  \\
      & HomoFormer~\cite{xiao2024homoformer} & 7.58  & 17.8  & 0.07  \\
      & MSRDNet~\cite{huang2025image} & 7.82  & 19.7  & 0.07  \\
      & StableShadow~\cite{xu2025detail} & 9.46  & 1329.8  & 0.72  \\
		\midrule
		\multirow{8}{*}{ZS}
		& Self-SGAN~\cite{jiang2023learning} & 7.55  & 3.7   & 91.73  \\
      & FLUX-Kontext~\cite{labs2025flux} & 13.44  & 8606.2  & 45.15  \\
      & Qwen-Image-Edit~\cite{wu2025qwen} & 19.54  & 15065.5  & 230.60  \\
      \cmidrule{2-5}
      & Ours $(T=10)$ & 7.79  & 949.6  & 2.59  \\
      & Ours $(T=20)$ & 7.34  & 949.6  & 4.17  \\
      & Ours $(T=30)$ & 7.22  & 949.6  & 5.88  \\
      & Ours $(T=40)$ & 7.16  & 949.6  & 7.47  \\
      & Ours $(T=50)$ & 7.15  & 949.6  & 8.99  \\
        \bottomrule
    \end{tabular}
%}
	\label{tab:compu}
\vspace{-0.1cm}
\end{table}

\begin{figure*}[!t]
    \centering
    \includegraphics[width=0.88\textwidth]{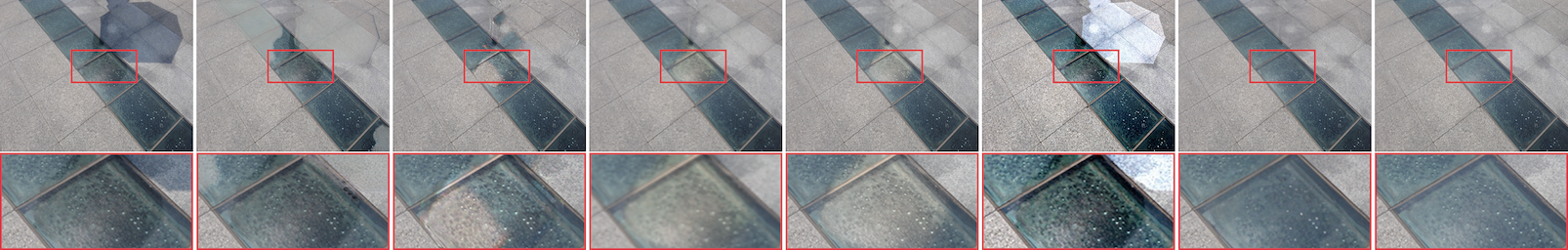}
    \\[0.15em]
    \includegraphics[width=0.88\textwidth]{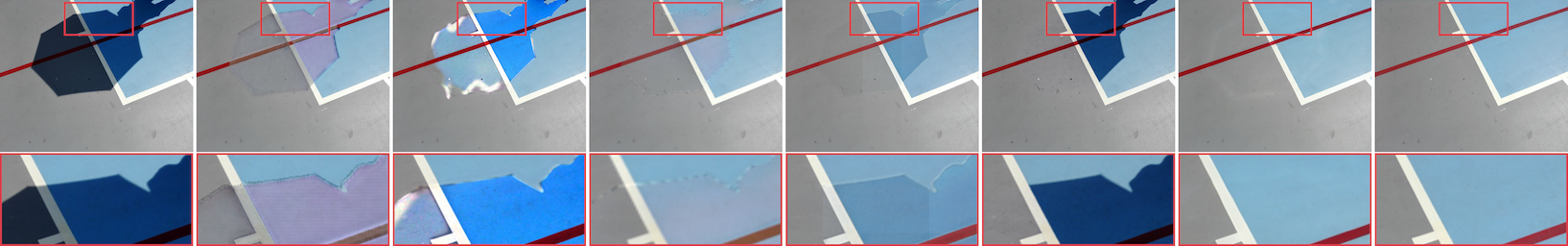}
    
    \footnotesize
    \begin{tabularx}{0.88\textwidth}{*{8}{>{\centering\arraybackslash}X}}
        Input & G2R-SNet & \mbox{Self-SGAN} & \mbox{\hspace{-0.6em}ShadowDiffusion} & HomoFormer & Kontext & Ours & GT
    \end{tabularx}
    
    \caption{Visual comparisons with state-of-the-art methods on the ISTD+ dataset.}
    \label{aistdfig}
\end{figure*}

\begin{figure*}[!t]
    \centering
    \includegraphics[width=0.88\textwidth]{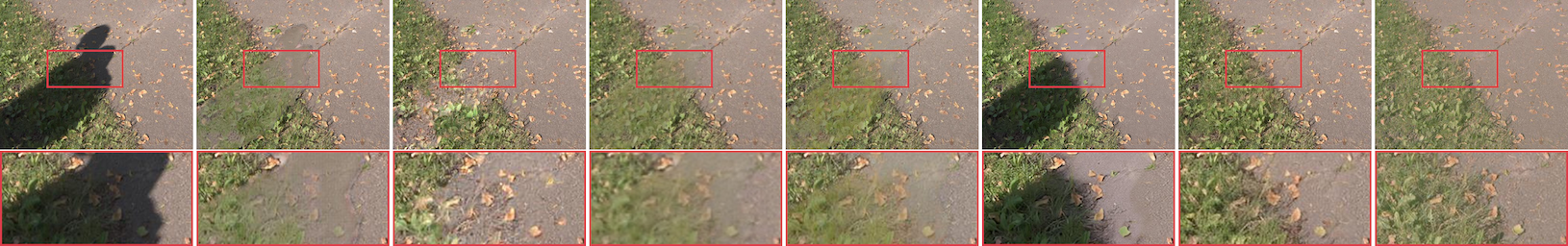}
    \\[0.15em]
    \includegraphics[width=0.88\textwidth]{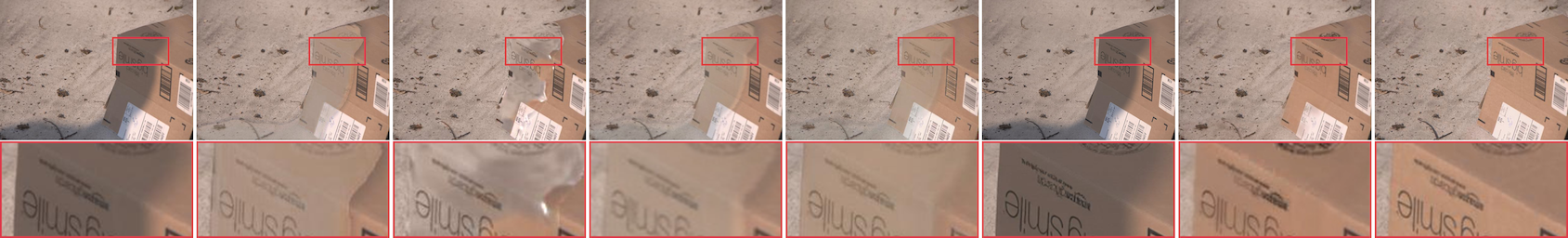}
    
    \footnotesize
    \begin{tabularx}{0.88\textwidth}{*{8}{>{\centering\arraybackslash}X}}
        Input & G2R-SNet & \mbox{Self-SGAN} & \mbox{\hspace{-0.6em}ShadowDiffusion} & \mbox{\hspace{-0.em}StableShadow} & Kontext & Ours & GT
    \end{tabularx}
    
    \caption{Visual comparisons with state-of-the-art methods on the UIUC+ dataset.}
    \label{uiucfig}
\end{figure*}

\subsubsection{Qualitative Comparison}
\begin{figure*}[!t]
    \centering
    \includegraphics[width=0.9\textwidth]{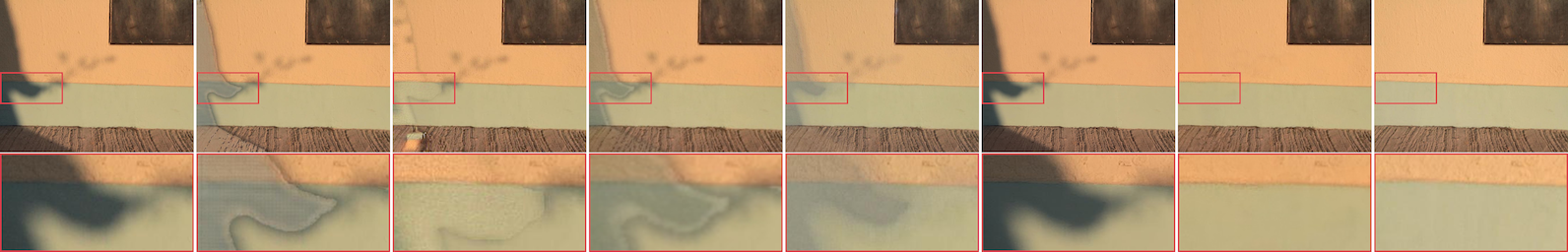}
    \\[0.15em]
    \includegraphics[width=0.9\textwidth]{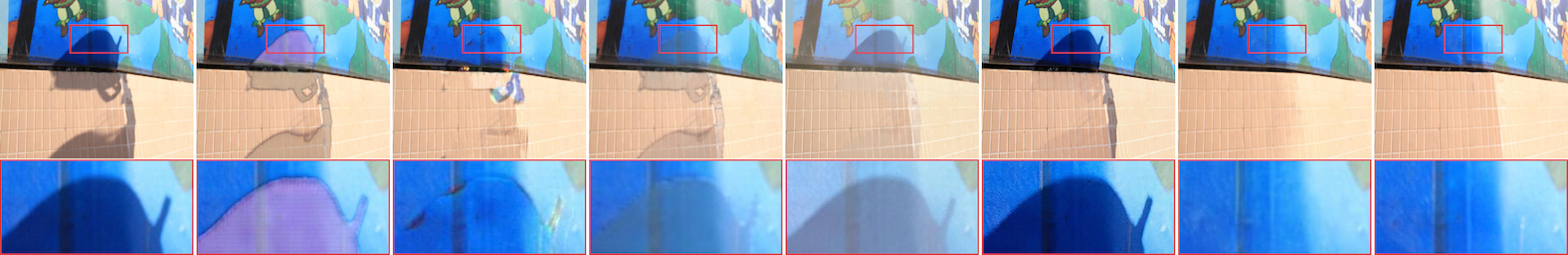}
    
    \footnotesize
    \begin{tabularx}{0.9\textwidth}{*{8}{>{\centering\arraybackslash}X}}
        Input & G2R-SNet & \mbox{Self-SGAN} & \mbox{\hspace{-0.6em}ShadowDiffusion} & \mbox{\hspace{-0.em}StableShadow} & Kontext & Ours & GT
    \end{tabularx}
    
    \caption{Visual comparisons with state-of-the-art methods on the SRD dataset.}
    \label{srdfig}
\end{figure*}

\begin{figure*}[!t]
    \centering
    \includegraphics[width=0.9\textwidth]{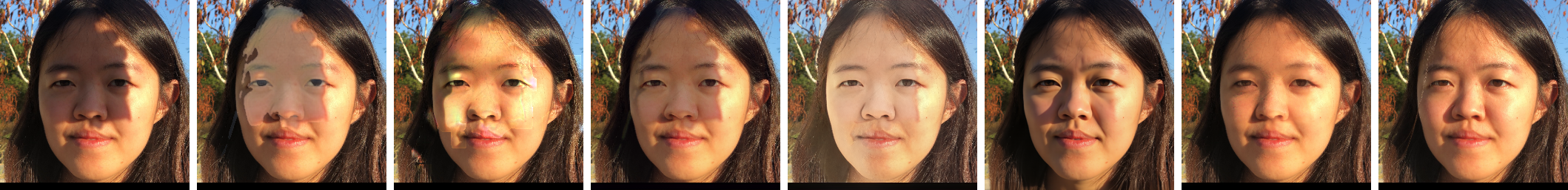}
    \\[0.15em]
    \includegraphics[width=0.9\textwidth]{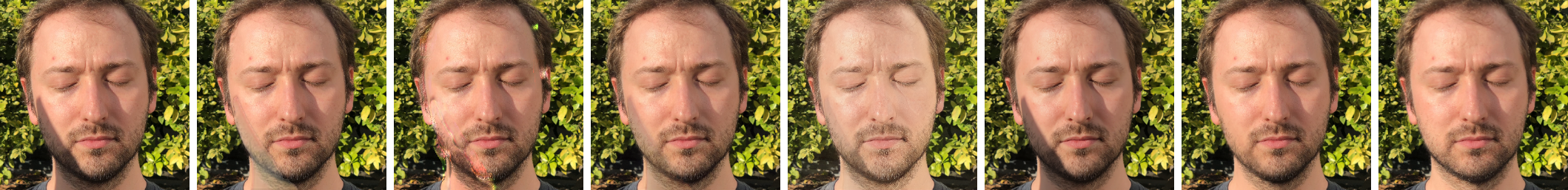}
    
    \footnotesize
    \begin{tabularx}{0.9\textwidth}{*{8}{>{\centering\arraybackslash}X}}
        Input & G2R-SNet & \mbox{Self-SGAN} & MSRDNet & \mbox{\hspace{-0.em}StableShadow} & Kontext & Ours & GT
    \end{tabularx}
    
    \caption{Visual comparisons with state-of-the-art methods on the PSM dataset.}
    \label{psmfig}
\end{figure*}

%\noindent
%{\bf Qualitative Comparison.}
Figs.~\ref{aistdfig}, \ref{uiucfig}, and \ref{srdfig} present visual comparisons of our method against competing methods on the ISTD+, UIUC+, and SRD datasets, respectively. Our method achieves superior visual results compared to existing methods. In particular, it more effectively restores illumination within shadow regions, ensuring seamless consistency with the surrounding non-shadow areas. Moreover, as shown in the first row of Fig.~\ref{uiucfig}, existing approaches that either train diffusion models from scratch (ShadowDiffusion) or fine-tune pretrained models (StableShadow) tend to generate shadow-region results that lack fine-grained details and exhibit limited perceptual realism. In contrast, our approach fully leverages the rich image priors encoded in the pretrained diffusion model, enabling the generation of more realistic shadow-free images. Existing image editing models (\eg, FLUX-Kontext) exhibit strong generative capabilities; however, they frequently fail to achieve thorough shadow removal. Fig.~\ref{psmfig} further shows visual comparisons on the PSM dataset. Our method effectively removes portrait shadows without introducing noticeable artifacts. Moreover, compared with StableShadow, our method better preserves facial color and produces more natural skin-tone consistency after shadow removal.

\subsection{Ablation Study}
We conduct ablation studies on the ISTD+ dataset to validate our main contributions.

\subsubsection{Effect of ITA}
%\noindent
%{\bf Effect of ITA.}
As shown in Table~\ref{tab:ablation_1} (b) and Fig.~\ref{fig:abla_1}, incorporating the ITA enhances overall illumination restoration within shadow regions, thereby improving shadow removal performance.

\subsubsection{Effect of SAMI and SHFI}
\begin{figure}[t]
    \centering
    \includegraphics[width=\linewidth]{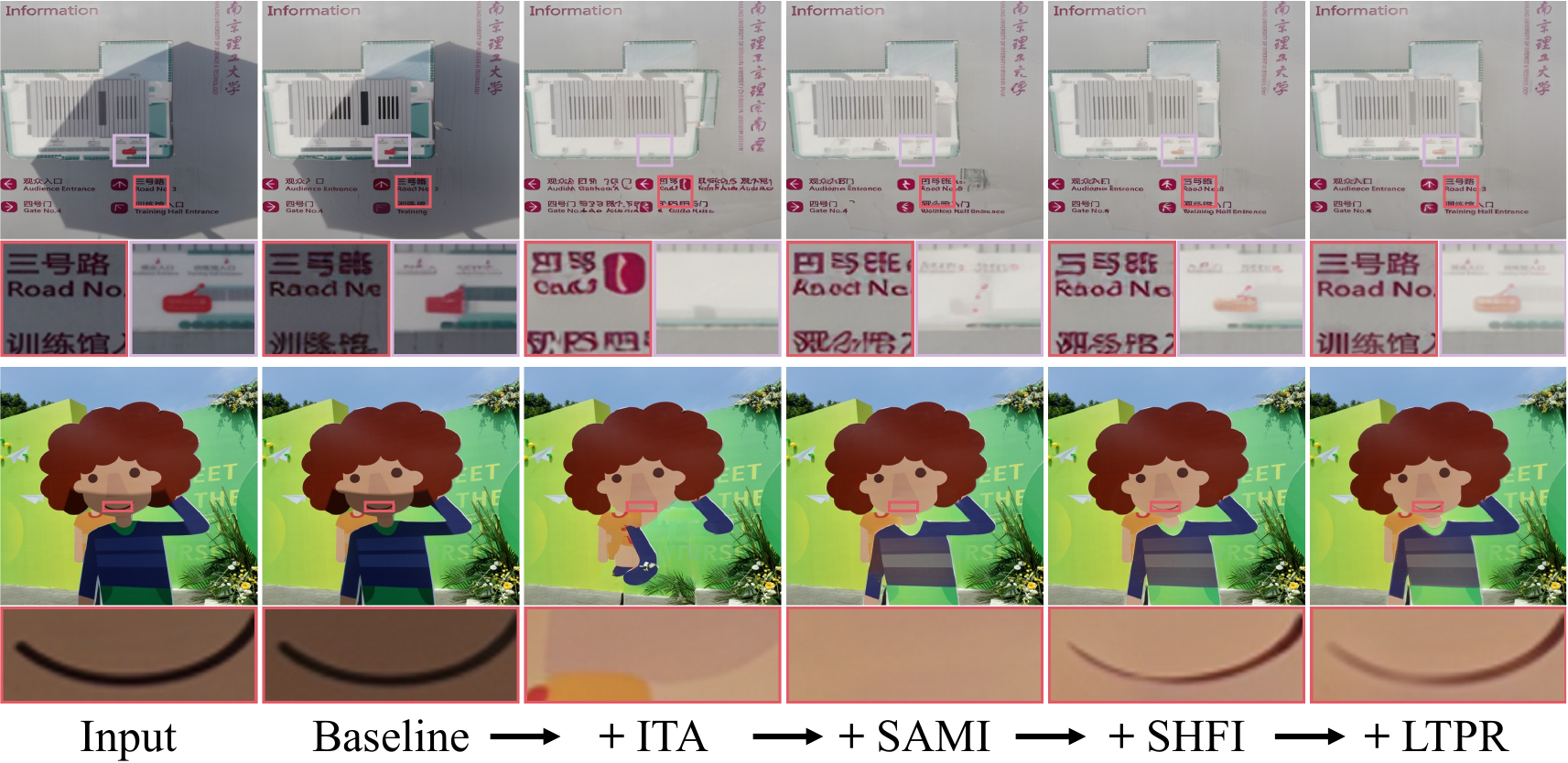}
    \caption{Visual effect of different components.}
    \label{fig:abla_1}
\end{figure}

%\noindent
%{\bf Effect of SAMI and SHFI.}
As illustrated in Fig.~\ref{fig:abla_1}, introducing SAMI helps preserve the overall structural integrity within shadow regions, while incorporating SHFI further retains fine-grained details. As indicated in Table~\ref{tab:ablation_1} (d) and (f), both SAMI and SHFI contribute positively to shadow removal performance. Notably, since SHFI primarily targets fine-scale details, its impact on performance is comparatively smaller.

\begin{table}[t]
\centering
\caption{Ablation studies of different components. \ding{52}\rotatebox[origin=c]{-9.2}{\kern-0.7em\ding{55}} indicates that the selective injection strategy is not employed in SAMI or SHFI.}
\footnotesize
\setlength{\tabcolsep}{0.5em}
% \vspace{-0.1cm}
\renewcommand{\arraystretch}{0.95}
%\adjustbox{width=\linewidth}{
    \begin{tabular}{c|cccc|ccc}
        \toprule
        	Model & ITA & SAMI & SHFI & LTPR & MAE$\downarrow$ & SSIM$\uparrow$ & PSNR$\uparrow$\\
       	\midrule
(a)   &       &       &       &       & 8.38  & 0.880  & 20.42  \\
(b)   & \ding{52} &       &       &       & 3.53  & 0.934  & 31.87  \\
(c)   & \ding{52} & \ding{52}\rotatebox[origin=c]{-9.2}{\kern-0.7em\ding{55}} &       &       & 4.81  & 0.927  & 25.50  \\
(d)   & \ding{52} & \ding{52} &       &       & 3.17  & 0.942  & 32.93  \\
(e)   & \ding{52} & \ding{52} & \ding{52}\rotatebox[origin=c]{-9.2}{\kern-0.7em\ding{55}} &       & 3.28  & 0.935  & 31.42  \\
(f)   & \ding{52} & \ding{52} & \ding{52} &       & 3.02  & 0.945  & 33.19  \\
\rowcolor[rgb]{ .902,  .902,  .902} (g)   & \ding{52} & \ding{52} & \ding{52} & \ding{52} & \textbf{2.90} & \textbf{0.960} & \textbf{33.80} \\
       	\bottomrule
    \end{tabular}
%}
\label{tab:ablation_1}
\end{table}

\begin{figure}[t]
    \centering
    \includegraphics[width=0.9\linewidth]{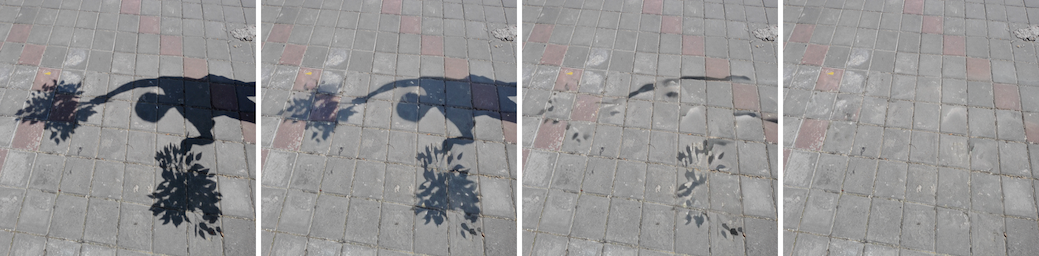}
    \\
    
    \footnotesize
    \begin{tabularx}{0.9\linewidth}{*{4}{>{\centering\arraybackslash}X}}
        Input & (c) & (e) & \mbox{(g) Full}
    \end{tabularx}
    
    \caption{Visual effect of the selective injection strategy. (c), (e), and (g) correspond to the models in Table~\ref{tab:ablation_1}. Specifically, (c) and (e) illustrate the results when the selective injection strategy is omitted in SAMI and SHFI, respectively.}
    \label{fig:abla_2}
\end{figure}

\subsubsection{Effect of Selective Injection Strategy}
%\noindent
%{\bf Effect of Selective Injection Strategy.}
In Table~\ref{tab:ablation_1}, model (c) represents a variant where all decoder self-attention maps are injected in SAMI, while model (e) represents a variant where the high-frequency components of the entire latent variables are injected in SHFI. Omitting the selective injection strategy in either SAMI or SHFI leads to a substantial performance drop. As shown in Fig.~\ref{fig:abla_2}, removing this strategy from SAMI results in noticeable residual shadows within shadow regions, whereas removing it in SHFI primarily causes shadow artifacts along shadow boundaries.

An analysis of the role of self-attention maps across different layers in SAMI is provided in Table~\ref{tab:abla_1}. As shown in Table~\ref{tab:abla_1} (h), (i), and (j), removing the injection of self-attention maps from any layer in the full model (g) leads to a decline in performance. Furthermore, as shown in Table~\ref{tab:abla_1} (k) and (l), injecting additional self-attention maps from the medium or high layers also results in performance degradation. This indicates the necessity of our selective design.

\begin{table}[!t]
\centering
\caption{Ablation study on attention map injection at different resolution layers.}
\footnotesize
\setlength{\tabcolsep}{0.5em}
% \vspace{-0.1cm}
\renewcommand{\arraystretch}{0.95}
%\adjustbox{width=0.5\textwidth}{
    \begin{tabular}{c|ccc|ccc}
        \toprule
        	Model & 16$\times$16 & 32$\times$32 & 64$\times$64 & MAE$\downarrow$ & SSIM$\uparrow$ & PSNR$\uparrow$ \\
       	\midrule
       	      (h)   & -     & 1   & 1     & 2.94  & 0.959  & \textbf{33.80} \\
(i)   & 1,2,3 & -     & 1     & 2.97  & 0.959  & \textbf{33.80} \\
(j)   & 1,2,3 & 1   & -     & 2.94  & 0.959  & 33.78  \\
\rowcolor[rgb]{ .902,  .902,  .902} (g)   & 1,2,3 & 1   & 1     & \textbf{2.90} & \textbf{0.960} & \textbf{33.80} \\
(k)   & 1,2,3 & 1,2,3 & 1     & 2.99  & 0.959  & 33.10  \\
(l)   & 1,2,3 & 1   & 1,2,3 & 25.83  & 0.400  & 16.77  \\
       	\bottomrule
    \end{tabular}
%}
\label{tab:abla_1}
\end{table}

\subsubsection{Effect of LTPR}
%\noindent
%{\bf Effect of LTPR.}
As illustrated in Fig.~\ref{fig:abla_1}, without applying LTPR, the images generated by SD exhibit noticeable local texture misalignment with the input shadow images. For instance, smaller text appears deformed and distorted. After incorporating LTPR, the original shapes of the text are well preserved. As shown in Table~\ref{tab:ablation_1} (g), LTPR leads to a significant improvement in quantitative shadow removal performance. This demonstrates that LTPR effectively retains the local texture of shadow image and enhances overall fidelity.

\begin{figure}[t]
%    \vspace{-0.5cm}
    \centering
    \includegraphics[width=0.8\linewidth]{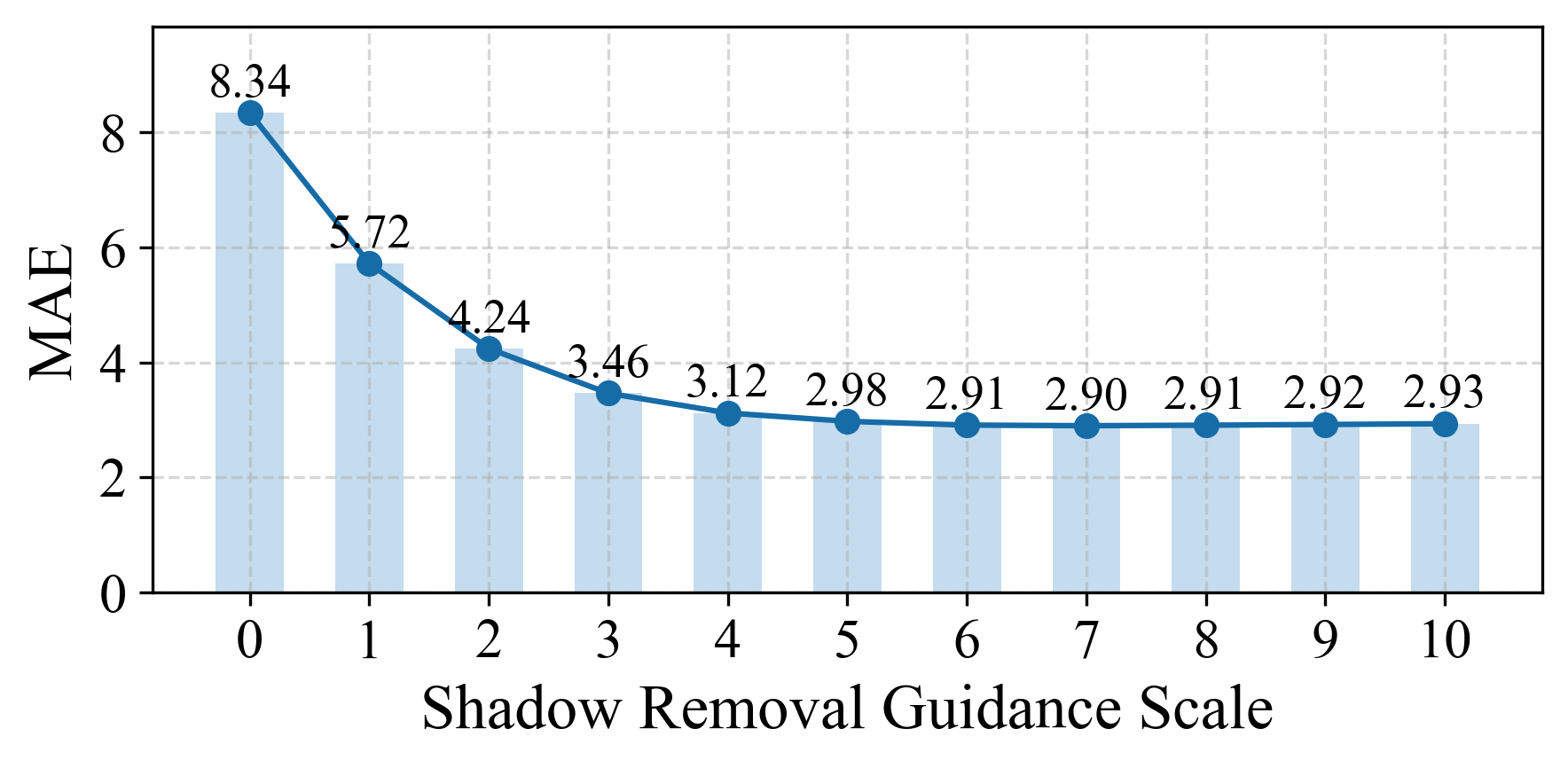}
    \vspace{-0.3cm}
    \caption{Ablation studies on shadow removal guidance scale.}
    \label{fig:rgs_bar_line}
%    \vspace{-1cm}
\end{figure}

\subsubsection{Effect of SRRG}
%\noindent
%{\bf Effect of Shadow Removal Redirect Guidance.}
To evaluate the effectiveness of the SRRG, we conduct ablation studies by varying the guidance scale $s$ in Eq.~(\ref{eq:srgs}). As illustrated in Fig.~\ref{fig:rgs_bar_line}, the shadow removal performance improves progressively as $s$ increases from $0$ to $6$. When $s$ exceeds $6$, the performance remains relatively stable. These results confirm that shadow removal redirect guidance plays a critical role in enhancing shadow removal performance. Therefore, $s$ is set to $7$ in our method. 
%Fig.~\ref{fig:srrg} further illustrates that as $s$ increases, shadow in the image are incrementally removed. 

\subsection{More Analysis}
\subsubsection{Robustness to Inaccurate Shadow Masks}
\begin{figure}[!t]
    \centering
    \raisebox{0.09\height}{\makebox[0.01\textwidth]{\rotatebox{90}{\makecell{\scriptsize (a) Soft Shadow}}}}
    \includegraphics[width=0.95\linewidth]{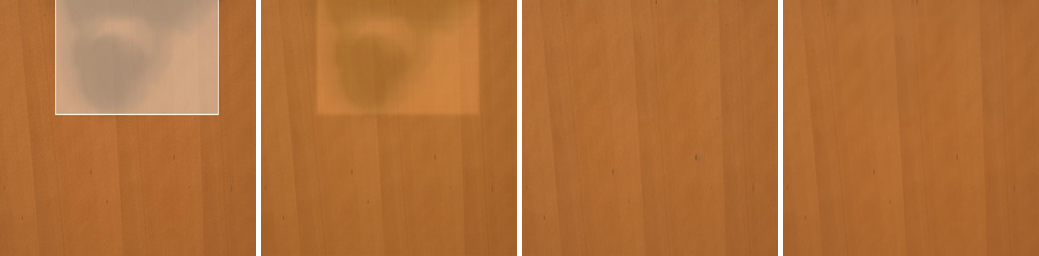}
    \\[0.2em]
    \raisebox{0.09\height}{\makebox[0.01\textwidth]{\rotatebox{90}{\makecell{\scriptsize (b) Hard Shadow}}}}
    \includegraphics[width=0.95\linewidth]{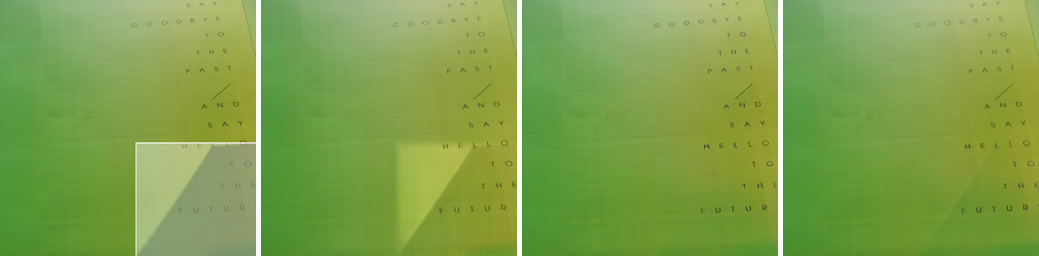}
    \\
    \footnotesize
    \makebox[0.01\textwidth]{}
	\begin{tabularx}{0.95\linewidth}{*{4}{>{\centering\arraybackslash}X}}
		\mbox{Input \& Mask} & Homoformer & \mbox{\hspace{-0.75em}Ours (ITA+SAMI)} & \mbox{Ours (Full)}
    \end{tabularx}
    \caption{Shadow removal performance under inaccurate shadow masks.}
    \label{fig:mask_box}
\end{figure}

%\noindent
%{\bf Robustness to Inaccurate Shadow Masks.}
To assess the robustness to inaccurate shadow masks, we adopt bounding-box masks that coarsely cover the shadow regions as input. The results are shown in Figs.~\ref{fig:mask_box}. Our method still achieves favorable shadow removal results under inaccurate mask guidance, while avoiding over-brightening regions that are incorrectly labeled as shadows. For hard shadows, the abrupt illumination transitions along shadow boundaries often degrade local details in these regions. Since SHFI and LTPR rely on the shadow boundary mask to restore fine details in degraded boundary regions, inaccurate shadow masks may cause the full model to produce boundary artifacts. By retaining only ITA and SAMI (excluding SHFI and LTPR), these artifacts can be effectively suppressed, providing a practical solution for achieving satisfactory shadow removal with coarse shadow masks.

\begin{figure}[!t]
    \centering
    \includegraphics[width=\linewidth]{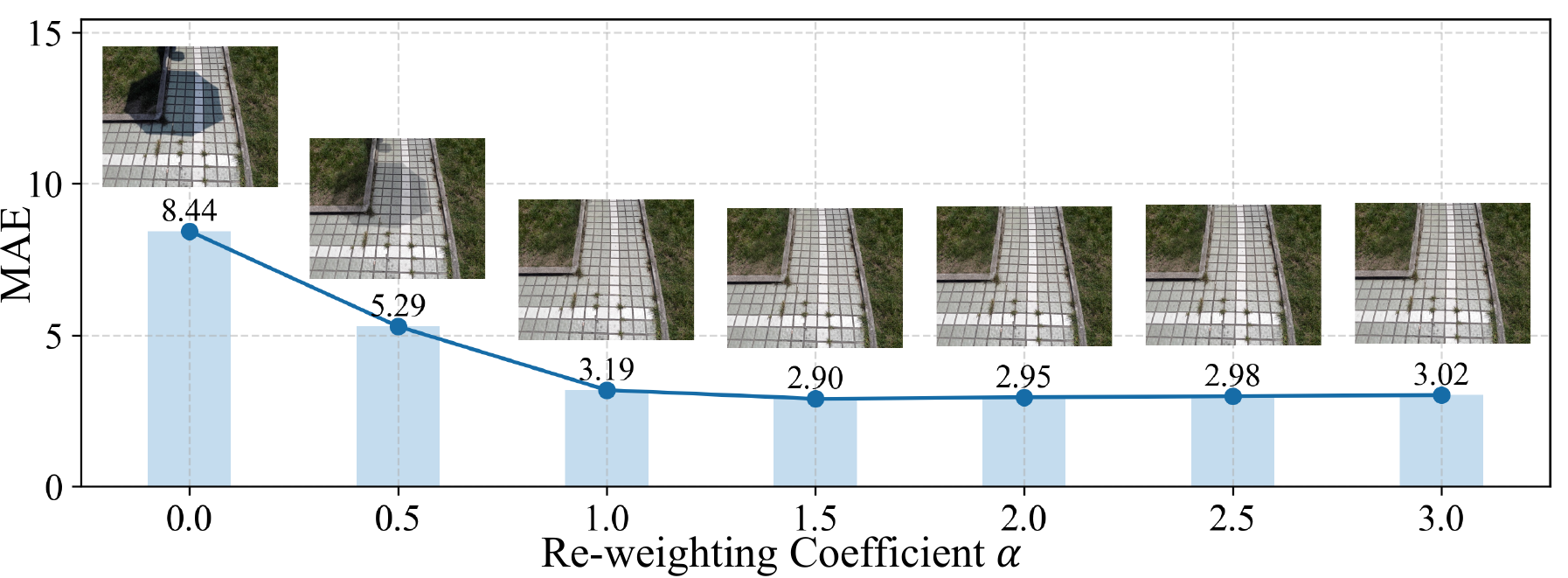}
    \caption{Analysis of re-weighting coefficient $\alpha$.}
    \label{fig:abla_alpha}
\end{figure}

\subsubsection{Parameter Sensitivity}
%\noindent
%{\bf Parameter Sensitivity.}
The analysis of the re-weighting coefficient $\alpha$ in ITA is illustrated in Fig.~\ref{fig:abla_alpha}. As $\alpha$ increases from $0$ to $1$, illumination transfer from non-shadow to shadow regions is progressively strengthened, leading to improved shadow removal performance. When $\alpha$ lies within the range of $1$ to $3$, our method achieves stable and consistently strong results. Notably, we adopt an identical set of hyperparameters across all four datasets, demonstrating that our approach exhibits low sensitivity to variations in data distribution.

\begin{figure}[!t]
    \centering
    \includegraphics[width=0.95\linewidth]{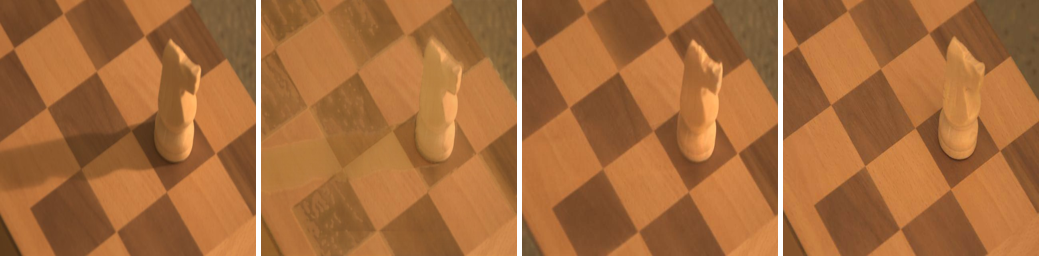}
    \footnotesize
    \begin{tabularx}{0.95\linewidth}{*{4}{>{\centering\arraybackslash}X}}
		Input & Homoformer & Ours & GT
    \end{tabularx}
    \caption{Failure cases of our method.}
    \label{fig:failure}
    \vspace{-0.1cm}
\end{figure}

\subsubsection{Limitations}
Fig.~\ref{fig:failure} shows a failure case of our method. Although our method can remove most cast shadows, it fails to completely eliminate the self-shadow on the chess piece. This is because self-shadows are closely entangled with object geometry and surface appearance, making them more difficult to remove than cast shadows. Improving the effectiveness of our method in such challenging cases will be an important direction for future work.

\section{Conclusion}
\label{sec:conclusion}
In this paper, we proposed FreeShadow, a training-free shadow removal framework that leverages pretrained diffusion priors without requiring paired data, model training, or inference-time optimization. By introducing illumination transfer attention, selective content preservation, and local texture-preserving relighting, FreeShadow effectively restores shadow-region illumination while maintaining structural consistency and fine-grained details. Extensive experiments on multiple benchmark datasets demonstrate that our method achieves competitive or superior performance against existing supervised, unsupervised, and zero-shot approaches, with strong generalization to diverse shadow scenes.
 
 % argument is your BibTeX string definitions and bibliography database(s)
%\bibliography{IEEEabrv,../bib/paper}

\bibliographystyle{IEEEtran}
\bibliography{main}

@article{guo2012paired,
  title={Paired regions for shadow detection and removal},
  author={Guo, Ruiqi and Dai, Qieyun and Hoiem, Derek},
  journal={IEEE TPAMI},
  volume={35},
  number={12},
  pages={2956--2967},
  year={2012},
}

@InProceedings{Zhong_2022_CVPR,
    author    = {Zhong, Yiqi and Liu, Xianming and Zhai, Deming and Jiang, Junjun and Ji, Xiangyang},
    title     = {Shadows Can Be Dangerous: Stealthy and Effective Physical-World Adversarial Attack by Natural Phenomenon},
    booktitle = {CVPR},
    year      = {2022},
    pages     = {15345-15354}
}

@inproceedings{qu2017deshadownet,
	title={Deshadownet: A multi-context embedding deep network for shadow removal},
	author={Qu, Liangqiong and Tian, Jiandong and He, Shengfeng and Tang, Yandong and Lau, Rynson WH},
	booktitle={CVPR},
	pages={4067--4075},
	year={2017}
}

@inproceedings{le2019shadow,
	title={Shadow removal via shadow image decomposition},
	author={Le, Hieu and Samaras, Dimitris},
	booktitle={ICCV},
	pages={8578--8587},
	year={2019}
}

@article{Guo_Huang_Liu_Cheng_Wen_2023, 
    title={ShadowFormer: Global Context Helps Shadow Removal}, 
    volume={37}, 
    number={1}, 
    journal={AAAI}, 
    author={Guo, Lanqing and Huang, Siyu and Liu, Ding and Cheng, Hao and Wen, Bihan}, 
    year={2023}, 
    pages={710-718}
}

@inproceedings{xu2025omnisr,
  title={Omnisr: Shadow removal under direct and indirect lighting},
  author={Xu, Jiamin and Li, Zelong and Zheng, Yuxin and Huang, Chenyu and Gu, Renshu and Xu, Weiwei and Xu, Gang},
  booktitle={AAAI},
  volume={39},
  number={8},
  pages={8887--8895},
  year={2025}
}

@inproceedings{hu2019mask,
  title={Mask-shadowgan: Learning to remove shadows from unpaired data},
  author={Hu, Xiaowei and Jiang, Yitong and Fu, Chi-Wing and Heng, Pheng-Ann},
  booktitle={ICCV},
  pages={2472--2481},
  year={2019}
}

@inproceedings{liu2021shadow,
  title={From shadow generation to shadow removal},
  author={Liu, Zhihao and Yin, Hui and Wu, Xinyi and Wu, Zhenyao and Mi, Yang and Wang, Song},
  booktitle={CVPR},
  pages={4927--4936},
  year={2021}
}

@article{jiang2023learning,
  title={Learning to remove shadows from a single image},
  author={Jiang, Hao and Zhang, Qing and Nie, Yongwei and Zhu, Lei and Zheng, Wei-Shi},
  journal={IJCV},
  volume={131},
  number={9},
  pages={2471--2488},
  year={2023},
  publisher={Springer}
}

@inproceedings{zheng2025shadow,
  title={When Shadow Removal Meets Intrinsic Image Decomposition: A Joint Learning Framework Using Unpaired Data},
  author={Zheng, Rongjia and Zhang, Qing and Nie, Yongwei and Zheng, Wei-Shi},
  booktitle={AAAI},
  volume={39},
  number={10},
  pages={10599--10607},
  year={2025}
}

@inproceedings{jin2021dc,
  title={Dc-shadownet: Single-image hard and soft shadow removal using unsupervised domain-classifier guided network},
  author={Jin, Yeying and Sharma, Aashish and Tan, Robby T},
  booktitle={ICCV},
  pages={5027--5036},
  year={2021}
}

@article{ho2020denoising,
  title={Denoising diffusion probabilistic models},
  author={Ho, Jonathan and Jain, Ajay and Abbeel, Pieter},
  journal={NeurIPS},
  volume={33},
  pages={6840--6851},
  year={2020}
}

@inproceedings{whang2022deblurring,
  title={Deblurring via stochastic refinement},
  author={Whang, Jay and Delbracio, Mauricio and Talebi, Hossein and Saharia, Chitwan and Dimakis, Alexandros G and Milanfar, Peyman},
  booktitle={CVPR},
  pages={16293--16303},
  year={2022}
}

@inproceedings{guo2023shadowdiffusion,
  title={Shadowdiffusion: When degradation prior meets diffusion model for shadow removal},
  author={Guo, Lanqing and Wang, Chong and Yang, Wenhan and Huang, Siyu and Wang, Yufei and Pfister, Hanspeter and Wen, Bihan},
  booktitle={CVPR},
  pages={14049--14058},
  year={2023}
}

@inproceedings{jin2024des3,
  title={Des3: Adaptive attention-driven self and soft shadow removal using vit similarity},
  author={Jin, Yeying and Ye, Wei and Yang, Wenhan and Yuan, Yuan and Tan, Robby T},
  booktitle={AAAI},
  volume={38},
  number={3},
  pages={2634--2642},
  year={2024}
}

@inproceedings{liu2024recasting,
  title={Recasting regional lighting for shadow removal},
  author={Liu, Yuhao and Ke, Zhanghan and Xu, Ke and Liu, Fang and Wang, Zhenwei and Lau, Rynson WH},
  booktitle={AAAI},
  volume={38},
  number={4},
  pages={3810--3818},
  year={2024}
}

@inproceedings{luo2025diff,
  title={Diff-shadow: Global-guided diffusion model for shadow removal},
  author={Luo, Jinting and Li, Ru and Jiang, Chengzhi and Zhang, Xiaoming and Han, Mingyan and Jiang, Ting and Fan, Haoqiang and Liu, Shuaicheng},
  booktitle={AAAI},
  volume={39},
  number={6},
  pages={5856--5864},
  year={2025}
}

@inproceedings{rombach2022high,
  title={High-resolution image synthesis with latent diffusion models},
  author={Rombach, Robin and Blattmann, Andreas and Lorenz, Dominik and Esser, Patrick and Ommer, Bj{\"o}rn},
  booktitle={CVPR},
  pages={10684--10695},
  year={2022}
}

@inproceedings{ramesh2021zero,
  title={Zero-shot text-to-image generation},
  author={Ramesh, Aditya and Pavlov, Mikhail and Goh, Gabriel and Gray, Scott and Voss, Chelsea and Radford, Alec and Chen, Mark and Sutskever, Ilya},
  booktitle={ICML},
  pages={8821--8831},
  year={2021},
  organization={Pmlr}
}

@inproceedings{guo2023boundary,
  title={Boundary-aware divide and conquer: A diffusion-based solution for unsupervised shadow removal},
  author={Guo, Lanqing and Wang, Chong and Yang, Wenhan and Wang, Yufei and Wen, Bihan},
  booktitle={ICCV},
  pages={13045--13054},
  year={2023}
}

@inproceedings{xu2025detail,
  title={Detail-Preserving Latent Diffusion for Stable Shadow Removal},
  author={Xu, Jiamin and Zheng, Yuxin and Li, Zelong and Wang, Chi and Gu, Renshu and Xu, Weiwei and Xu, Gang},
  booktitle={CVPR},
  pages={7592--7602},
  year={2025}
}

@article{huang2025image,
  title={Image shadow removal via multi-scale deep Retinex decomposition},
  author={Huang, Yan and Lu, Xinchang and Quan, Yuhui and Xu, Yong and Ji, Hui},
  journal={PR},
  volume={159},
  pages={111126},
  year={2025}
}

@article{wang2024exploiting,
  title={Exploiting diffusion prior for real-world image super-resolution},
  author={Wang, Jianyi and Yue, Zongsheng and Zhou, Shangchen and Chan, Kelvin CK and Loy, Chen Change},
  journal={IJCV},
  volume={132},
  number={12},
  pages={5929--5949},
  year={2024},
  publisher={Springer}
}

@inproceedings{huang2025zero,
  title={Zero-Shot Low-Light Image Enhancement via Latent Diffusion Models},
  author={Huang, Yan and Liao, Xiaoshan and Liang, Jinxiu and Quan, Yuhui and Shi, Boxin and Xu, Yong},
  booktitle={AAAI},
  volume={39},
  number={4},
  pages={3815--3823},
  year={2025}
}

@inproceedings{songdenoising,
  title={Denoising Diffusion Implicit Models},
  author={Jiaming Song and Chenlin Meng and Stefano Ermon},
  booktitle={ICLR},
  year={2021},
}

@inproceedings{tumanyan2023plug,
  title={Plug-and-play diffusion features for text-driven image-to-image translation},
  author={Tumanyan, Narek and Geyer, Michal and Bagon, Shai and Dekel, Tali},
  booktitle={CVPR},
  pages={1921--1930},
  year={2023}
}

@InProceedings{xu2025stylessp,
    author={Xu, Ruojun and Xi, Weijie and Wang, XiaoDi and Mao, Yongbo and Cheng, Zach},
    title={StyleSSP: Sampling StartPoint Enhancement for Training-free Diffusion-based Method for Style Transfer},
    booktitle={CVPR},
    year={2025},
    pages={18260-18269}
}

@inproceedings{mei2024latent,
  title={Latent feature-guided diffusion models for shadow removal},
  author={Mei, Kangfu and Figueroa, Luis and Lin, Zhe and Ding, Zhihong and Cohen, Scott and Patel, Vishal M},
  booktitle={WACV},
  pages={4313--4322},
  year={2024}
}

@article{liu2023decoupled,
  title={A decoupled multi-task network for shadow removal},
  author={Liu, Jiawei and Wang, Qiang and Fan, Huijie and Li, Wentao and Qu, Liangqiong and Tang, Yandong},
  journal={IEEE TMM},
  volume={25},
  pages={9449--9463},
  year={2023}
}

@inproceedings{xiao2024homoformer,
  title={Homoformer: Homogenized transformer for image shadow removal},
  author={Xiao, Jie and Fu, Xueyang and Zhu, Yurui and Li, Dong and Huang, Jie and Zhu, Kai and Zha, Zheng-Jun},
  booktitle={CVPR},
  pages={25617--25626},
  year={2024}
}

@article{guo2024single,
  title={Single-image shadow removal using deep learning: A comprehensive survey},
  author={Guo, Laniqng and Wang, Chong and Wang, Yufei and Yu, Yi and Huang, Siyu and Yang, Wenhan and Kot, Alex C and Wen, Bihan},
  journal={arXiv preprint arXiv:2407.08865},
  year={2024}
}

@article{schaerf2025training,
  title={Training-Free Style and Content Transfer by Leveraging U-Net Skip Connections in Stable Diffusion 2.},
  author={Schaerf, Ludovica and Alfarano, Andrea and Silvestri, Fabrizio and Impett, Leonardo},
  journal={arXiv preprint arXiv:2501.14524},
  year={2025}
}

@article{yang2012shadow,
  title={Shadow removal using bilateral filtering},
  author={Yang, Qingxiong and Tan, Kar-Han and Ahuja, Narendra},
  journal={IEEE TIP},
  volume={21},
  number={10},
  pages={4361--4368},
  year={2012},
  publisher={IEEE}
}

@article{yoon2024generative,
  title={Generative portrait shadow removal},
  author={Yoon, Jae Shin and Shu, Zhixin and Ren, Mengwei and Zhang, Cecilia and Hold-Geoffroy, Yannick and Singh, Krishna Kumar and Zhang, He},
  journal={ACM TOG},
  volume={43},
  number={6},
  pages={1--13},
  year={2024},
  publisher={ACM New York, NY, USA}
}

@article{zhang2020portrait,
  title={Portrait shadow manipulation},
  author={Zhang, Xuaner and Barron, Jonathan T and Tsai, Yun-Ta and Pandey, Rohit and Zhang, Xiuming and Ng, Ren and Jacobs, David E},
  journal={ACM TOG},
  volume={39},
  number={4},
  pages={78--1},
  year={2020},
  publisher={ACM New York, NY, USA}
}

@article{labs2025flux,
  title={FLUX. 1 Kontext: Flow Matching for In-Context Image Generation and Editing in Latent Space},
  author={Labs, Black Forest and Batifol, Stephen and Blattmann, Andreas and Boesel, Frederic and Consul, Saksham and Diagne, Cyril and Dockhorn, Tim and English, Jack and English, Zion and Esser, Patrick and others},
  journal={arXiv preprint arXiv:2506.15742},
  year={2025}
}

@article{wu2025qwen,
  title={Qwen-image technical report},
  author={Wu, Chenfei and Li, Jiahao and Zhou, Jingren and Lin, Junyang and Gao, Kaiyuan and Yan, Kun and Yin, Sheng-ming and Bai, Shuai and Xu, Xiao and Chen, Yilei and others},
  journal={arXiv preprint arXiv:2508.02324},
  year={2025}
}

@article{podell2023sdxl,
  title={Sdxl: Improving latent diffusion models for high-resolution image synthesis},
  author={Podell, Dustin and English, Zion and Lacey, Kyle and Blattmann, Andreas and Dockhorn, Tim and M{\"u}ller, Jonas and Penna, Joe and Rombach, Robin},
  journal={arXiv preprint arXiv:2307.01952},
  year={2023}
}

@article{niu2022boundary,
  title={A boundary-aware network for shadow removal},
  author={Niu, Kunpeng and Liu, Yanli and Wu, Enhua and Xing, Guanyu},
  journal={IEEE TMM},
  volume={25},
  pages={6782--6793},
  year={2022},
  publisher={IEEE}
}

@inproceedings{le2020shadow,
  title={From shadow segmentation to shadow removal},
  author={Le, Hieu and Samaras, Dimitris},
  booktitle={ECCV},
  pages={264--281},
  year={2020},
  organization={Springer}
}

@article{shao2021hyper,
  title={Hyper RPCA: joint maximum correntropy criterion and Laplacian scale mixture modeling on-the-fly for moving object detection},
  author={Shao, Zerui and Pu, Yifei and Zhou, Jiliu and Wen, Bihan and Zhang, Yi},
  journal={IEEE TMM},
  volume={25},
  pages={112--125},
  year={2021}
}

@article{zhang2019decoupled,
  title={Decoupled spatial neural attention for weakly supervised semantic segmentation},
  author={Zhang, Tianyi and Lin, Guosheng and Cai, Jianfei and Shen, Tong and Shen, Chunhua and Kot, Alex C},
  journal={IEEE TMM},
  volume={21},
  number={11},
  pages={2930--2941},
  year={2019}
}

@article{xie2023consistency,
  title={Consistency preservation and feature entropy regularization for gan based face editing},
  author={Xie, Weicheng and Lu, Wenya and Peng, Zhibin and Shen, Linlin},
  journal={IEEE TMM},
  volume={25},
  pages={8892--8905},
  year={2023},
  publisher={IEEE}
}

@article{hu2025icdsr,
  title={ICDSR: Integrated Conditional Diffusion Model for Single Image Super-Resolution},
  author={Hu, Cong and Wei, Xiao-Zhong and Wu, Xiao-Jun},
  journal={IEEE TMM},
  year={2026},
  publisher={IEEE},
  volume={28},
  pages={1302-1313},
}

@article{zhang2024mmginpainting,
  title={Mmginpainting: Multi-modality guided image inpainting based on diffusion models},
  author={Zhang, Cong and Yang, Wenxia and Li, Xin and Han, Huan},
  journal={IEEE TMM},
  volume={26},
  pages={8811--8823},
  year={2024},
  publisher={IEEE}
}

@article{sultana2020unsupervised,
  title={Unsupervised moving object detection in complex scenes using adversarial regularizations},
  author={Sultana, Maryam and Mahmood, Arif and Jung, Soon Ki},
  journal={IEEE TMM},
  volume={23},
  pages={2005--2018},
  year={2020}
}

\vfill

\end{document}